\documentclass[10pt,twocolumn,letterpaper]{article}

\usepackage[pagenumbers]{cvpr} 

\usepackage{acro}
\usepackage{float}
\usepackage{graphicx}
\usepackage{subcaption}
\usepackage{amsmath}
\usepackage{multirow}
\usepackage{gensymb}
\usepackage{amsfonts}
\usepackage{pifont}
\usepackage{bm}
\usepackage{enumitem}

\renewcommand{\emph}[1]{\textit{{#1}}}

\renewcommand{\paragraph}[1]{\vspace{0.75em}  \noindent \textbf{#1} }

\usepackage{subcaption}
\usepackage{mathtools}
\usepackage{booktabs} %
\usepackage{mathrsfs}
\usepackage{enumitem}

\usepackage{forloop}

\newcommand{\defcal}[1]{\expandafter\newcommand\csname 
	c#1\endcsname{{\mathcal{#1}}}}
\newcommand{\defbb}[1]{\expandafter\newcommand\csname 
	b#1\endcsname{{\mathbb{#1}}}}
\newcommand{\defbf}[1]{\expandafter\newcommand\csname 
	bf#1\endcsname{{\bm{#1}}}}
\newcounter{calBbCounter}
\forLoop{1}{26}{calBbCounter}{
	\edef\letter{\Alph{calBbCounter}}
	\expandafter\defcal\letter
	\expandafter\defbb\letter
	\expandafter\defbf\letter
}
\forLoop{1}{26}{calBbCounter}{
	\edef\letter{\alph{calBbCounter}}
	\expandafter\defbf\letter
}

\usepackage{pifont}
\usepackage[pagebackref,breaklinks,colorlinks]{hyperref}

\usepackage{acro}
\usepackage{bigdelim}
\usepackage{multibib}
\usepackage{makecell}

\usepackage{color, colortbl}
\usepackage[accsupp]{axessibility}
\DeclareAcronym{ood}{
  short = OOD ,
  long = out-of-distribution
}

\definecolor{Gray}{gray}{0.9}

\newcites{supp}{References}%

\def\cvprPaperID{3296} 
\def\confName{CVPR}
\def\confYear{2022}

\newcommand{\thedataset}{SHIFT\xspace}

\begin{document}

\title{\thedataset: A Synthetic Driving Dataset for Continuous Multi-Task \\Domain Adaptation}

\author{Tao Sun$^{1*}$ \quad Mattia Segu$^{1}$\thanks{Equal contribution.} \quad Janis Postels$^1$ \quad Yuxuan Wang$^1$ \quad \\
Luc Van Gool$^1$ \quad Bernt Schiele$^2$ \quad Federico Tombari$^{3,4}$ \quad Fisher Yu$^1$\\[0.4cm]
$^1$ETH Z\"urich \quad $^2$MPI Informatics \quad $^3$Google \quad $^4$Technical University of Munich \\
{\tt\small \{taosun47, segum, jpostels, yuxuwang\}@ethz.ch}\\ {\tt\small vangool@vision.ee.ethz.ch, schiele@mpi-inf.mpg.de, tombari@in.tum.de, i@yf.io}}

\maketitle

\begin{abstract}
Adapting to a continuously evolving environment is a safety-critical challenge inevitably faced by all autonomous driving systems.
Existing image and video driving datasets, however, fall short of capturing the mutable nature of the real world.
%
In this paper, we introduce the largest multi-task synthetic dataset for autonomous driving, \thedataset{}. 
It presents discrete and continuous shifts in cloudiness, rain and fog intensity, time of day, and vehicle and pedestrian density. 
Featuring a comprehensive sensor suite and annotations for several mainstream perception tasks, \thedataset{} allows investigating the degradation of a perception system performance at increasing levels of domain shift, fostering the development of continuous adaptation strategies to mitigate this problem and assess model robustness and generality.
%
%
Our dataset and benchmark toolkit are publicly available at \href{https://www.vis.xyz/shift}{\texttt{www.vis.xyz/shift}}. 
\end{abstract}


    

\section{Introduction}

Recent years have witnessed the remarkable progress of perception systems for autonomous driving.
%
%
Betting on the role that autonomous driving will serve for society,
industry
and academia have joined forces to collect and release several large-scale driving datasets, raising hopes for a forthcoming successful deployment of self-driving cars.

Large-scale driving datasets have 
played a pivotal role in the prosperity of perception algorithms and provide a playground for different techniques to compete and thrive on multiple tasks.
However, while the algorithm accuracy surges, progress in terms of generalization to unforeseen environmental conditions has been underwhelming~\cite{dai2018dark,michaelis2019benchmarking}.

\begin{figure}[!t]
    \centering
    \includegraphics[width=\linewidth]{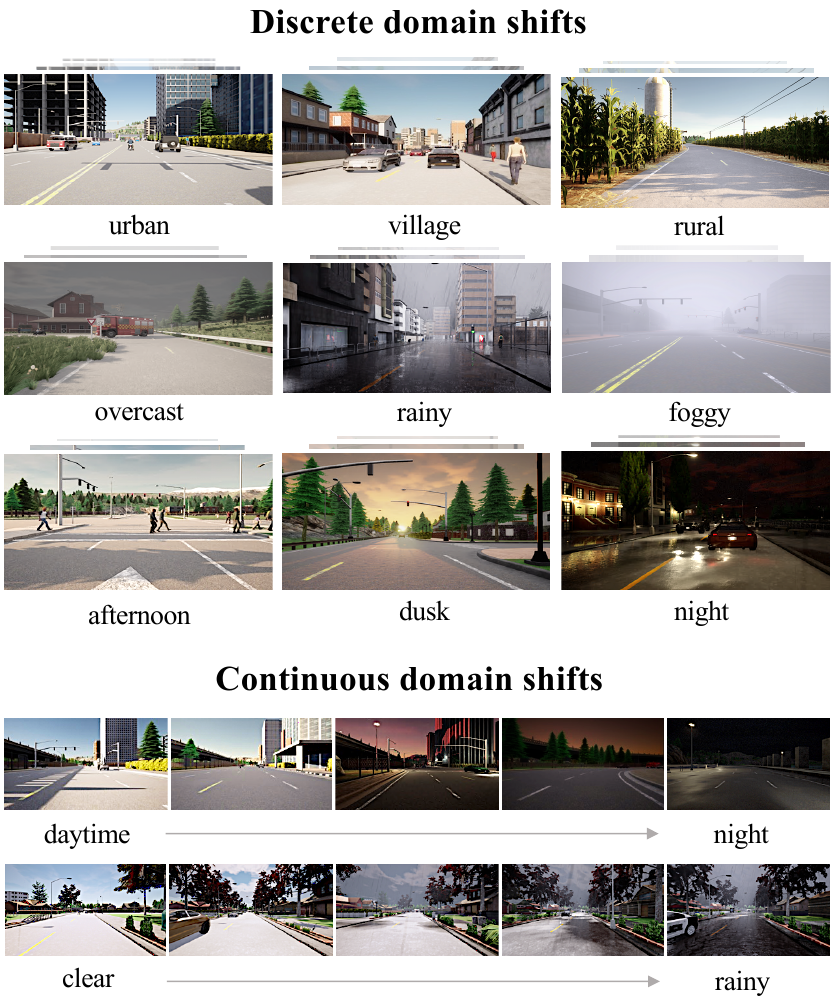}
    \caption{\thedataset{} provides: (a) \textbf{discrete domain shifts}, a set of sequences each collected using different domain parameters and initial states;
    (b) \textbf{continuous domain shifts}, a set of sequences where domain parameters change continuously during driving.
    \label{fig:teaser}}
\end{figure}

To achieve full autonomy, self-driving cars must adapt to new environments and identify life-threatening failure cases to promptly prevent crashes.
Examples of domain shifts affecting driving are changes in weather and lighting conditions, scenery, and behavior, appearance, and quantity of agents
on the road.
Domain shift~\cite{ben2010theory} is a well-known problem for learning algorithms, causing unforeseeable performance drops under conditions different from the training ones. 
Techniques to prevent, counteract or assess its impact have been developed in the form of domain generalization~\cite{khosla2012undoing,muandet2013domain,tobin2017domain,volpi2018generalizing}, domain adaptation~\cite{wang2018deep,ganin2015unsupervised,long2015learning,tzeng2017adversarial}, uncertainty estimation~\cite{gal2016dropout,lakshminarayanan2017simple,postels2019sampling,loquercio2020general} and \ac{ood} detection~\cite{yang2021generalized,ruff2021unifying,hendrycks2016baseline,postels2020hidden}.
However, such approaches are typically deployed and tested on toy datasets~\cite{saenko2010adapting,venkateswara2017deep,li2017deeper} or synthetically corrupted ones~\cite{hendrycks2019robustness}. 
Although there are preliminary attempts at providing driving datasets with different domains~\cite{bdd100k,caesar2020nuscenes,dai2018dark,pitropov2021canadian,sakaridis2018semantic,tung2017raincouver,sakaridis2021acdc,ros2016synthia}, each only covers a limited amount of perception tasks (\eg{} only semantic segmentation~\cite{sakaridis2018semantic,sakaridis2021acdc}) and a narrow selection of domain shift directions (\eg{} only rain~\cite{tung2017raincouver} or snow~\cite{pitropov2021canadian}). 
Consequently, current solutions to domain shift cannot undergo scrutiny in controlled autonomous driving scenarios, making it difficult to verify their safety without risking real-world car crashes.
%

%
Given their short length, sequences from existing driving datasets are captured under approximately stationary conditions, and only \textit{discrete shifts} are witnessed among sets of sequences presenting different homogeneous conditions from one set to another (\eg{} clear weather and rainy).
However, nothing in this world is constant except change and becoming.
\textit{Continuous shifts} - the intra-sequence shifts from one domain into another - are a certainty in the real world, where a sunny day can rapidly turn into a rainy one, or a quiet road can quickly become busy.
Moreover, continuous distributional shift has recently been shown to represent a critical challenge for current learning systems~\cite{postels2021practicality}.

An adequate dataset design is thus needed to quantify and address domain shift both at discrete and continuous levels.
%
Consequently, we set the goal of overcoming the outdated paradigm of previous driving datasets and introduce \thedataset{}, a new synthetic dataset capturing the continuously evolving nature of the real world through realistic discrete and continuous shifts along safety-critical environmental directions: time of day, cloudiness, rain, fog strength, and vehicle and pedestrian density.
Collected in the CARLA simulator~\cite{Dosovitskiy17}, \thedataset{} includes a comprehensive sensor suite and covers the most important perception tasks. Counting 4,800+ sequences captured from a multi-view sensor suite in 8 different locations, our dataset supports 13 perception tasks for multi-task driving systems: semantic/instance segmentation, monocular/stereo depth regression, 2D/3D object detection, 2D/3D multiple object tracking (MOT), optical flow estimation, point cloud registration, visual odometry, trajectory forecasting and human pose estimation.

Our dataset aims to foster research in several under-explored fields related to the generality and reliability of perception systems for autonomous driving, \eg{} domain generalization, domain adaptation, and uncertainty estimation.
Moreover, by collecting incremental discrete shifts from one domain to another, we hope to foster research in the field of continual learning~\cite{wang2020tent,volpi2021continual,hacohen2019power} for autonomous driving, so far only studied on discrete levels of synthetic corruptions~\cite{hendrycks2019robustness} of traditional image classification datasets~\cite{deng2009imagenet,krizhevsky2014cifar}.
%
%
Finally, by collecting sequences with realistic intra-sequence continuous domain shifts, we provide the first driving dataset allowing research on continuous test-time learning and adaptation~\cite{wang2020tent,sun2020test,tonioni2019learning,tonioni2019real,poggi2021continual}.  
%

The main contributions of this work are:
\begin{itemize}[leftmargin=1.25em]
\vspace{-0.5em}
\setlength\itemsep{0em}
    \item We introduce \thedataset{}, a multi-task driving dataset featuring the most important perception tasks under a variety of conditions and with a comprehensive sensor setup.
    To the best of our knowledge, it is the largest synthetic dataset for autonomous driving and provides the most inclusive set of annotations and conditions.
    %
    %
    \item Using \thedataset{}, we analyze the importance of modeling discrete and continuous domain shifts, and demonstrate new findings on different adaptation and uncertainty estimation methods under continuous shifts.
\end{itemize}

\section{Related Work}
%
\begin{table*}[!t]
\centering
\footnotesize
\setlength{\tabcolsep}{5pt}
\begin{tabular}{clccccccccccc}
\toprule
& \multirow{2}{*}{\textbf{Dataset}}                                                     & \multirow{2}{*}{\textbf{Cities}} & \textbf{Tracking} & \textbf{\textbf{Max length for}} & \textbf{\textbf{Domain}} & \multicolumn{6}{c}{\textbf{Annotated frames for}}                                                                           \\ \cline{7-13} 
                         &                                                              &                                  & \textbf{sequences}                                    & \textbf{sequence}                & \textbf{shift$^\dagger$}       & Seg.                 & 2D Det.              & 3D Det.              & MOT           & Depth    & Flow        & Pose$^\diamondsuit$         \\ \midrule
\parbox[t]{2mm}{\multirow{10}{*}{\rotatebox[origin=c]{90}{\textbf{Real-world}}}} & KITTI~\cite{geiger2013vision}                                   & 1                                & 22                                  &  106 sec                                                & no                                            & 200                  & 15k                  & 15k                  & 15k                  &    93k      & 389         & -          \\
& CamVid~\cite{brostow2009semantic}                               & 4                                & -                                   & -                                                  & no                                            & 700                  & -                    & -                    & -                    & -        & -        & -            \\
 & Cityscapes~\cite{cordts2016cityscapes}                          & 27                               & -                                   & -                                                  & no                                            & 25k                  & 25k                    & 25k                    & -                    & -        & -           & -          \\
& Cityscapes-C$^\ddagger$ ~\cite{michaelis2019benchmarking}                   & 27                               & -                                   & -                                                 & discrete                                      & 25k                  & 25k                    & 25k                    & -                    & -        & -              & -       \\
& H3D~\cite{patil2019h3d}                                         & 4                                & 160                                 &    20 sec                                                & discrete                                      & -                    & -                    & 27k                  & 27k                  & -        & -                & -    \\
& HD1K~\cite{kondermann2016hci}                                   & 1                                & -                                   & -                                                  & discrete                                      & -                    & -                    & -                    & -                    & -        & 1k                 & -  \\
& A*3D~\cite{pham2020a3d}                                         & 1                                & -                                   & -                                                 & discrete                                      & -                    & -                    & 39k                  & -                    & -        & -                 & -   \\
& nuScenes~\cite{caesar2020nuscenes}                              & 2                                & 1,000                                & 20 sec                                            & discrete                                      & -                    & -                    & 40k                  & 40k                  & -        & -                 & -   \\
& Waymo Open~\cite{sun2020scalability}                            & 3                                & 1,150                                & 20 sec                                            & discrete                                      & -                    & 200k                 & 230k                 & 230k                 & -        & -          & 230k          \\
& BDD100K~\cite{bdd100k}                                          & -$^\mathsection$                         & 2,000                                & 40 sec                                           & discrete                                      & 10k                  & 100k                 & -                    & 318k                 & -        & -             & -       \\ \midrule
\parbox[t]{2mm}{\multirow{6}{*}{\rotatebox[origin=c]{90}{\textbf{Synthetic}}}} & SYNTHIA~\cite{ros2016synthia}                                   & 3                                & -                                   & -                                                  & discrete                                      & 9,000                 & 200k                 & 200k                 & -                    & -        & -               & -      \\ 
& GTA-V~\cite{richter2016playing}                                 & 1                                & -                                   & -                                                 & no                                            & 25k                  & -                    & -                    & -                    & -        & -          & -           \\
& VIPER~\cite{Richter_2017} & 1    &  184  &  10 min & discrete & 320k & 320k & - & 320k & - & 320k & - \\
& AIODrive~\cite{Weng2020_AIODrive}                              & 8                                & 100                                 & 100 sec                                            & discrete                                      & 100k                 & 100k                 & 100k                 & 100k                 & 100k     & -          & -          \\
\cmidrule{2-13}
& \textbf{\thedataset (ours)}  & {8}                            & 4,850                                & 33 min                                          & 
\makecell{discrete +\\continuous}                     & {2.5M}             & {2.5M}             & {2.5M}             & {2.5M}             & {2.5M} & {2.5M} & {2.5M} \\
\bottomrule
\end{tabular}
\vspace{-0.5em}
\caption{Comparison of size and supported tasks of existing driving datasets. \thedataset{} is the largest synthetic dataset and, most notably, the only dataset providing realistic continuous domain shifts, diverse annotations, and longer annotated sequences. $^\dagger$ indicates whether the dataset presents domain annotations. $^\ddagger$ artificially corrupted. $^\mathsection$ multiple cities; exact number not known. $^\diamondsuit$ key points for human pose.}
\label{tab:datasets}
\vspace{-0.5em}
\end{table*}
During the past decade, a large variety of realistic and synthetic driving datasets emerged, providing a playground for researchers to develop novel algorithms. Domain shift is a common threat to the performance and safety of learning-based methods.

We here introduce the most notable driving datasets and the techniques to mitigate the domain shift effect.
%
For an overview of the
current driving datasets, refer to \autoref{tab:datasets}.

\paragraph{Real-world driving datasets} typically focus on a specific subset of perception tasks due to the high cost of data collection and annotation.  
After almost a decade of development, the pioneering real-world dataset KITTI~\cite{geiger2013vision} supports almost all the perception tasks for autonomous driving, including semantic / instance segmentation, depth estimation, 2D and 3D object detection and tracking, optical flow, scene flow, and visual odometry. However, its small scale represents an obvious problem and its diversity is severely limited compared to modern large-scale datasets. 
CamVid~\cite{brostow2009semantic}, Cityscapes~\cite{cordts2016cityscapes}, and Mapillary~\cite{neuhold2017mapillary} are image-based driving datasets for segmentation, A*3D~\cite{pham20203d} for 3D object detection, and HD1K~\cite{kondermann2016hci} for optical flow estimation.
Recently, many large-scale datasets, \textit{e.g.}, BDD100K~\cite{bdd100k}, Waymo Open~\cite{sun2020scalability}, H3D~\cite{patil2019h3d}, and nuScenes~\cite{caesar2020nuscenes}, have been released with multi-task annotations, although mainly focusing on object detection and tracking. 
Our dataset offers a complete set of annotations for all the frames, comprehensive of all the most important perception tasks supported by other datasets, and enabling multi-task learning on a broader range of tasks and conditions.

\paragraph{Synthetic driving datasets} are collected using graphic engines and physical simulators. SYNTHIA~\cite{ros2016synthia} contains images and segmentation annotations generated by its simulator.
AIODrive~\cite{Weng2020_AIODrive} is produced using CARLA Simulator with multiple sensor support, focusing on high-density long-range LiDAR sets. 
Compared to ours, these datasets present sequences of limited length and are restricted to discrete domain labels (\autoref{tab:datasets}).
Further, video games have also been used for data generation. GTA-V~\cite{richter2016playing,hu2021monocular}  provides images and segmentation masks captured from a popular game. VIPER~\cite{Richter_2017} extends GTA-V by providing optical flow masks and discrete environmental labels. However, low-level control of video game engines is hardly accessible, impeding fine-grained environmental control and the collection of continuous shifts.

\paragraph{Adverse conditions datasets} support the evaluation of robustness under different \ac{ood} conditions.
A recent work~\cite{malinin2021shifts} collects meteorological and air temperature measurements under discrete real-world shifts.
Image-based datasets, \eg{}  CIFAR10/100-C~\cite{michaelis2019benchmarking}, ImageNet-R~\cite{hendrycks2021many} and Cityscapes-C~\cite{hendrycks2019robustness}, have been generated by applying artificial corruptions such as blurring, additive Gaussian noise and addition of specific patterns on the original dataset. Though carefully designed, such ad-hoc corruptions cannot fully represent the challenges presented by visual shifts in the real world.
%
%
To this end, recent driving datasets~\cite{pham2020a3d, bdd100k, caesar2020nuscenes, sun2020scalability,mao2021one} provide manually labeled tags for various weather conditions, scene categories, and day periods. 
However, each only covers a limited amount of perception tasks (see \autoref{tab:datasets}) and a narrow selection of domain shift directions. 
Moreover, ad-hoc datasets have been collected for specific underrepresented domains, \eg{} rain~\cite{tung2017raincouver,jin2021raidar}, fog~\cite{tarel2012vision,sakaridis2018semantic,sakaridis2018model}, night~\cite{dai2018dark}, snow~\cite{pitropov2021canadian}.
However, domain tags remain coarse-grained and only certain tasks and domain shift directions are supported.
Recently, the ACDC dataset~\cite{sakaridis2021acdc} has been proposed, featuring images evenly distributed between fog, nighttime, rain, and snow. However, it supports only semantic segmentation.
Interestingly, the India Driving Dataset~\cite{varma2019idd} is the only dataset to provide extremely busy roads as adverse conditions.
Overall, BDD100K~\cite{bdd100k} is the large-scale real-world dataset presenting the largest diversity of perception driving tasks and discrete domain labels for the time of day and weather conditions.
%
%
For this reason, we use it as a reference to validate empirical observations drawn from our dataset.
Nevertheless, compared to our dataset, BDD100K only provides annotated images from single cameras, does not provide 3D bounding boxes and optical flow annotations, distribution of domains is highly imbalanced and the domain is stationary within each sequence.
In contrast, our dataset provides a full sensor suite, annotations for multiple tasks, balanced domain distribution and sets of sequences with continuously changing time of day, weather conditions (cloudiness, rain and fog strength), and vehicle and pedestrians density.

\begin{figure*}[!t]
    \centering
    \vspace{-1em}
    \includegraphics[width=0.99\linewidth]{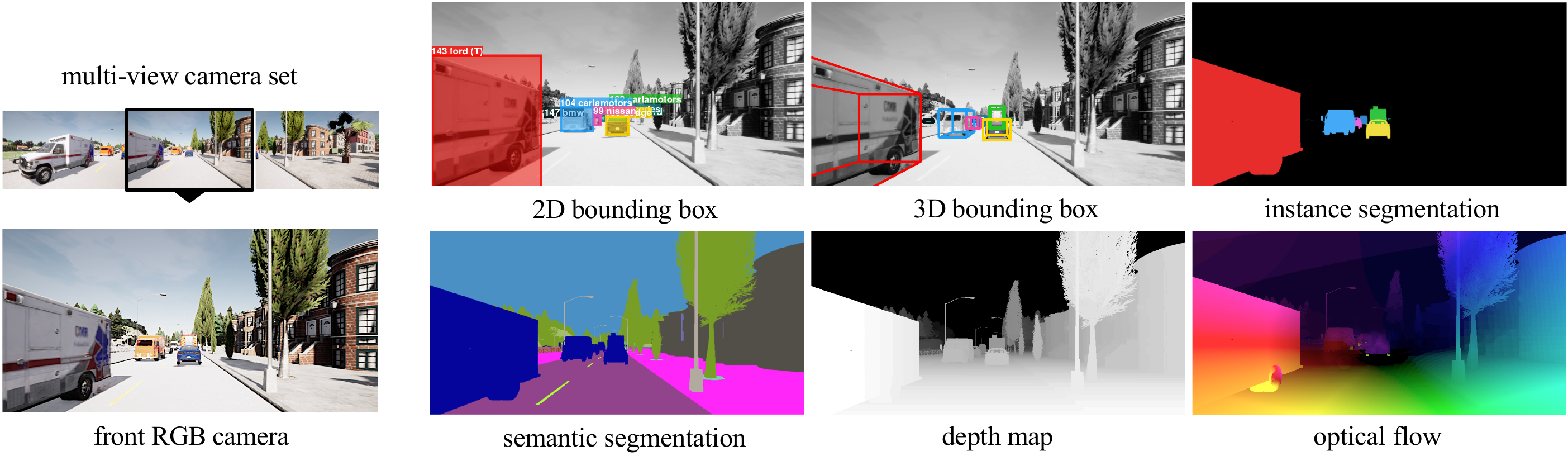}
    \caption{The annotation set of the RGB camera in our dataset. Each frame is associated with annotations of 2D/3D bounding boxes with tracking identities (visualized by different colors), semantic/instance segmentation, depth map and optical flow label. \label{figure:ann}}
\end{figure*}

\paragraph{Unsupervised domain adaptation (UDA)} means simultaneously learning on a labeled source and an unlabeled target domain to find transferable features across domains. UDA is mainly achieved via feature-space alignment~\cite{quinonero2008covariate, sun2017correlation}, domain-consistent regularization~\cite{ganin2015unsupervised, ganin2016domain,hoffman2018cycada} and minimization of surrogate functions of domain gaps~\cite{vu2019advent, saito2019semi}. The discrete shifts provided in our dataset can be directly used for training and evaluating UDA approaches.

\paragraph{Continual domain adaptation} aims at performing consecutive discrete adaptation steps from one domain to multiple others.
Incremental domain adaptation (IncDA) is a subset of continual DA that requires the source data and assumes availability of intermediate domains where domain shifts occur gradually~\cite{wang2020tent,volpi2021continual,lao2020continuous}, allowing to minimize the gap between adaptation steps and performing adaptation from the source to the final target domain more effectively than with direct UDA.
Providing different strengths of variations along natural axes, our dataset is suitable for IncDA.

\paragraph{Continuous test-time adaptation} (ContinuousTTA) assumes that gradual domain shifts occur within the same test sequence, and adaptation is performed at test time on the incoming data stream. 
ContinuousTTA is a suitable choice for any scenario where a model is required to adapt on the go to a shifting domain and no large labeled or unlabeled collection of data from the target domain is available in advance. 
Recent works~\cite{wang2020tent,mummadi2021test,sun2020test} show the efficiency of TTA when applied to artificial corruptions in the image-based datasets ImageNet-C/-R~\cite{hendrycks2019robustness,hendrycks2021many}. 
The continuously shifting video sequences in our dataset provide instead realistic domain shift along natural directions, facilitating the development of ContinuousTTA methods transferable to the real world.

\paragraph{Uncertainty Estimation} is a fundamental task for safety-critical vision applications. Quantifying the confidence about a model's prediction allows avoiding dangerous failures in autonomous driving.
However, current uncertainty estimation techniques~\cite{lakshminarayanan2017simple,gal2016dropout,liu2020simple,postels2020hidden} mainly focus on classification on toy datasets~\cite{krizhevsky2014cifar,lecun1998mnist}, while recent work~\cite{postels2021practicality} has observed poor calibration, \ie{} uncertainty uncorrelated with prediction's error, when such techniques are extended to more difficult datasets~\cite{hendrycks2019benchmarking} and tasks under distributional shift.
We hope that the domain shifts and multiple tasks supported in \thedataset{} will enable the study of uncertainty estimation methods on a wide variety of tasks for autonomous driving and their calibration under distributional shift.

\section{The \thedataset{} Dataset}
We provide a driving dataset with a comprehensive sensor suite (\autoref{ssec:sensors}) and a rich set of annotations (\autoref{ssec:annotations}), supporting multiple image- and video-based perception and forecasting tasks against environmental changes. 
We detail our design choices regarding domain shifts in \autoref{ssec:dataset_design}.

\subsection{Sensor Suite} \label{ssec:sensors}
We collect the data through a comprehensive sensor suite. 
Our sensor suite features 11 different sensors, including a multi-view RGB camera set with 5 cameras, a stereo RGB camera set, an optical flow sensor, a depth camera, a GNSS sensor, and an IMU. All the cameras have a field-of-view of $90\degree$ and resolution of $1280 \times 800$ pixel.
Moreover, we provide point clouds captured by a 128-channel LiDAR sensor.
%
All sensors are synchronized and captured at a $10 \mathrm{Hz}$ rate. 
We follow the Scalabel~\cite{scalabel} format and right-hand coordinate systems for storing all the annotations.
%
%
More details are in the Appendix.


\subsection{Annotations} \label{ssec:annotations}
We provide annotations for multiple mainstream perception tasks in autonomous driving, including 2D/3D bounding box trajectories,
instance/semantic segmentation, optical flow and dense depth.
Unlike real-world datasets, whose annotations are often limited to a group of keyframes due to prohibitive labeling cost, we offer full annotations for each frame in the sequences.
More details are in the Appendix.
%


\begin{figure*}[t!]
    \centering
    \includegraphics[width=0.95\linewidth]{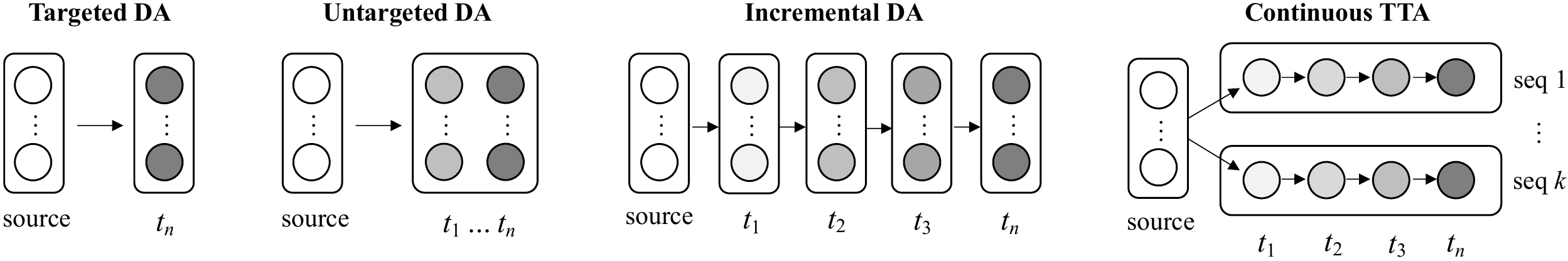}
    \caption{We evaluate four adaptation strategies: targeted domain adaptation (Targeted DA), untargeted domain adaptation (Untargeted DA), incremental domain adaptation (Incremental DA) and continuous test-time adaptation (Continuous TTA). The dots in the same row represent frames from the same sequence; their grayscale marks the degree of domain shift (white dots = source, dark gray dots = target.) }
    \label{fig:adaptation_strategies}
\end{figure*}

\begin{figure}[t]
    \centering
    \includegraphics[width=1.02\linewidth]{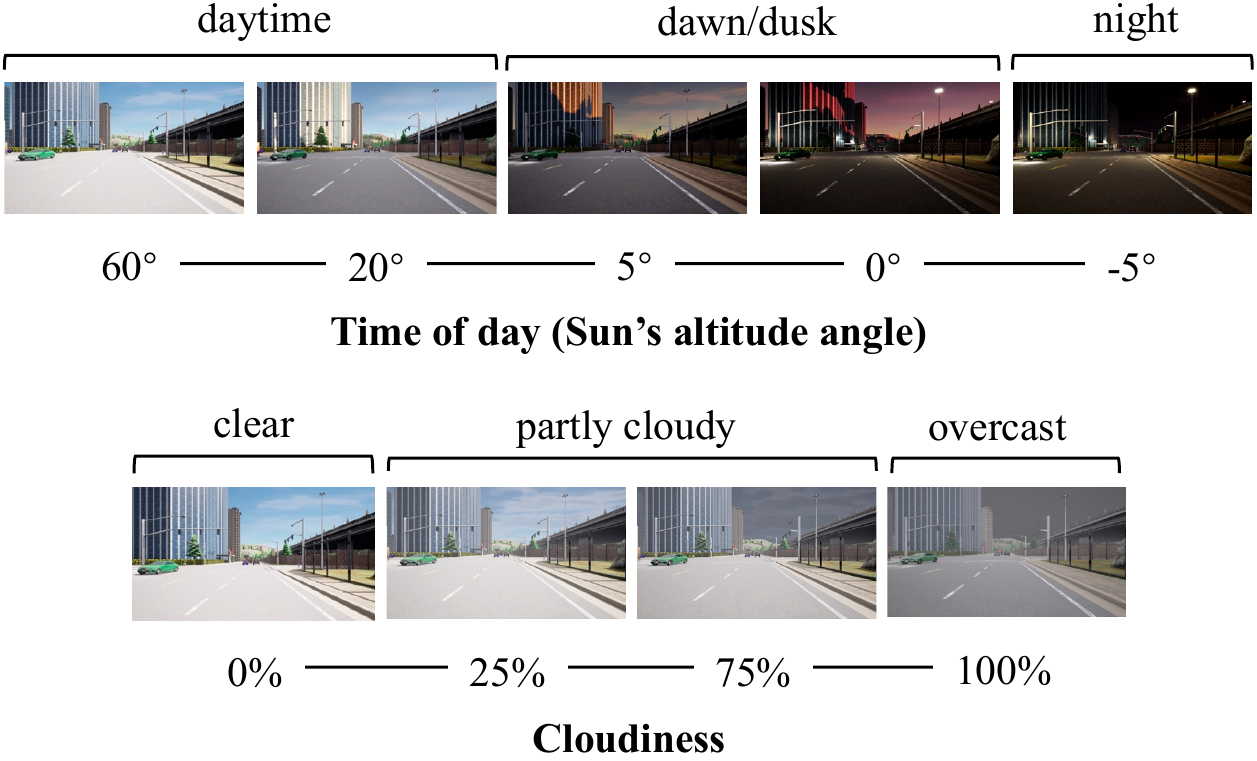}
    \vspace{-2em}
    \caption{Examples of the two-level structure for domain labels. Each discrete label (tag above images) corresponds to an interval of continuous labels (\textit{i.e.}, severity, axis below images). }
    \label{fig:cont_vs_}
    \vspace{-1em}
\end{figure}

\subsection{Dataset Design} \label{ssec:dataset_design}
Given their short sequence length, existing driving datasets are captured under approximately stationary conditions, and only discrete shifts are witnessed among sets of sequences presenting different homogeneous conditions (\eg{} clear weather and rainy).
We set the goal of overcoming the outdated paradigm of previous driving datasets and introduce \thedataset{}, a new synthetic dataset capturing the continuously evolving nature of the real world through realistic discrete and continuous shifts along safety-critical environmental directions: time of day, cloudiness, rain, fog strength, and vehicle and pedestrian density.
We collect 5,250 sequences, of which 4,250 contain stationary 
environmental conditions, \ie{} inter-sequence domain shift. Each sequence is composed of 500 frames collected at 10 Hz, equivalent to 50 seconds of driving time. The remaining 600 sequences have continuously shifting conditions, \ie{} inter-sequence domain shift. 
Totalling 70+ hours of driving and 2,500,000 annotated frames, \thedataset{} is the largest synthetic driving dataset available.
%


\paragraph{Domain shift types.} We consider the most-frequent real-world environmental changes. \thedataset{} provides domain shifts in (a) weather conditions, including cloudiness, rain, and fog intensity, (b) time of day, (c) the density of vehicles and pedestrians, and (d) camera orientation. 


\paragraph{Domain shifts level.} To facilitate research on domain adaptation in different scenarios, \thedataset{} provides two levels of domain shifts, namely discrete domain shifts and continuous domain shifts. 
The \textit{discrete} set contains 4,250 sequences generated with fixed environmental parameters and random initial states. We group these sequences into different domains, according to their severity. \autoref{fig:cont_vs_} shows grouping examples. All possible domain combinations are uniformly distributed across all sequences. 
The \textit{continuous} set contains additional 600 sequences with continuous domain variations. In particular, each sequence presents a gradual shift from one domain to another, where the shift happens through the intermediate domains that would naturally occur in the real world. In total, we collect 500 sequences of a basic 40 seconds length (1x), 80 sequences 10x longer than the basic length, and 20 100x longer. Each set is uniformly divided among the following shifts, each of which also loops back to the source domain: day $\xrightarrow{}$ night,  clear $\xrightarrow{}$ rain, clear $\xrightarrow{}$ foggy, clear $\xrightarrow{}$ overcast. Given a domain shift direction, \eg{} day to night, all other domain parameters are uniformly distributed across all sequences. 
%
%
Different sequence lengths allow analyzing the impact of domain shift speed on continuous TTA strategies (\autoref{ssec:exp_continuous}).






\begin{figure*}[t!]
    \centering
    \vspace{-1em}
    \subfloat[Object detection]{\includegraphics[width=0.46\linewidth]{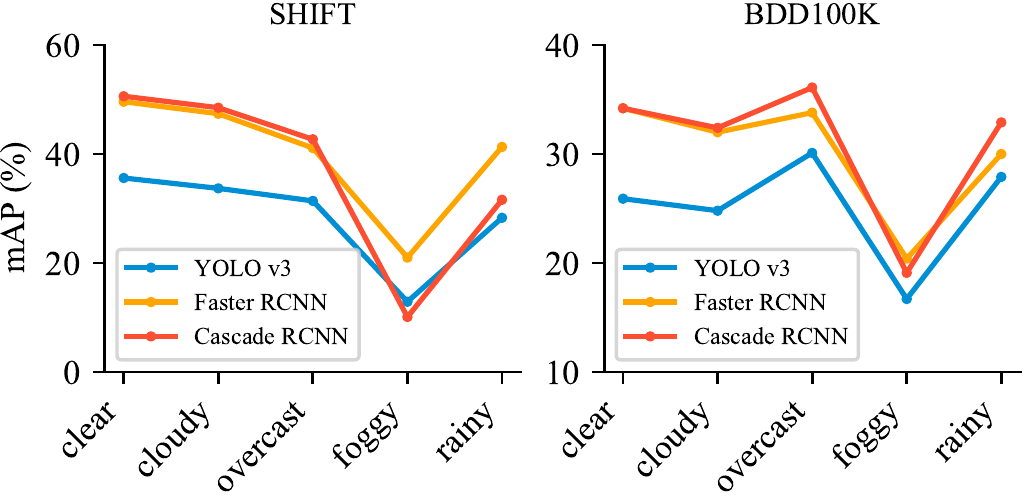}} \hspace{8mm}
    \subfloat[Semantic segmentation]{\includegraphics[width=0.46\linewidth]{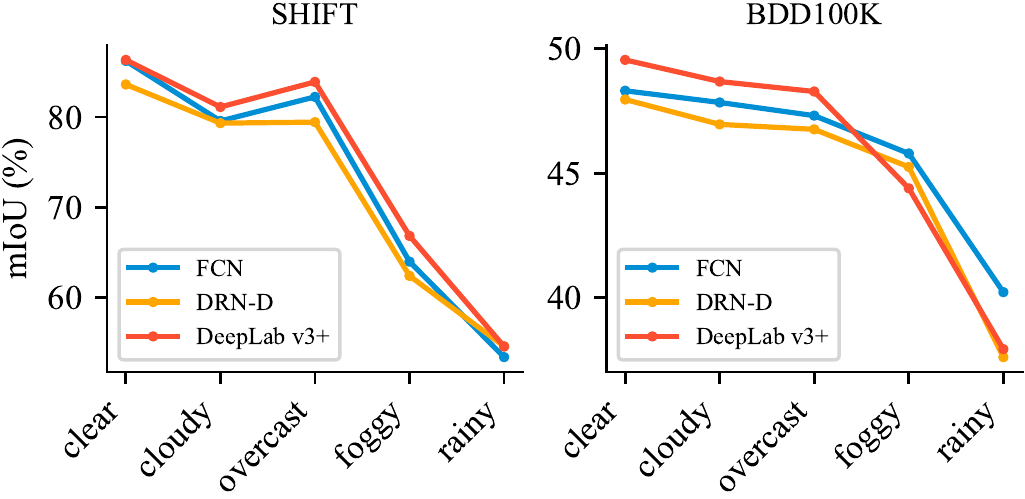}}
    \vspace{-0.5em}
    \caption{Performance degradation for different object detection (left) and semantic segmentation (right) methods under different weather conditions. Every model is trained under clear weather conditions and tested on other domains. \thedataset shows a similar trend as BDD100K. }
    \label{fig:trend}
\end{figure*}

\begin{table*}[t!]
\centering
\footnotesize
\setlength{\tabcolsep}{5pt}
\begin{tabular}{lllccccccc}
\toprule
 \textbf{Task}    & \textbf{Method}  &  \textbf{Metric}     &  \underline{clear-daytime}     & partly cloudy & overcast & foggy & rainy  & dawn/dusk & night        \\ \midrule
 Semantic segmentation & DRN-D~\cite{yu2017dilated} & mIoU (\%) $\uparrow$ & 83.6 & 79.3 & 79.4 & 62.4 & 54.6 & 60.8 & 42.8  \\
Instance segmentation & Mask R-CNN~\cite{he2017mask} & mAP (\%) $\uparrow$ &  39.3 & 39.4 & 34.0 & 18.7 &  35.0 & 30.7 & 13.1 \\
Object detection & Faster R-CNN~\cite{cai2018cascade} & mAP (\%) $\uparrow$& 46.9 & 47.4 & 41.1 & 21.0 & 41.3 & 37.3 & 15.4  \\
MOT & QDTrack~\cite{pang2021quasi} & MOTA (\%) $\uparrow$ & 56.2 & 53.4 & 46.2 & 25.0 & 41.9 & 44.7 & 16.5 \\
Mono. depth estimation & AdaBins-UNet~\cite{bhat2021adabins} & SILog $\downarrow$ & 9.6 & 10.0 &  8.9 & 12.0 & 10.3 & 19.7 & 27.9 \\
Optical flow estimation & RAFT~\cite{teed2020raft} & EPE (px) $\downarrow$ &  2.26 & 2.01 & 2.35 & 2.60 &  2.43 &  4.17 & 8.85 \\
\bottomrule
\end{tabular}
\caption{Performance degradation on \thedataset of different methods for different perception tasks under discrete domain shifts. Training domain is underlined. The test domains are weather variations in daytime (partly cloudy, overcast, foggy, rainy) and time of day variations in clear weather (dawn/dusk, night). $\uparrow$ ($\downarrow$): the higher (lower) the better.} \label{tab:ood_performance}
\vspace{-3mm}
\end{table*}

\begin{table}[t]
\centering
\footnotesize
\setlength{\tabcolsep}{6pt}

\begin{tabular}{lccc}
\toprule
\textbf{Scenario}             & \multicolumn{1}{c}{{Baseline}} & {Targeted DA} & \multicolumn{1}{c}{{Incremental DA}} \\ \midrule
\underline{daytime} $\rightarrow$ night & 42.8                           & 45.3                                  & \textbf{47.3}                        \\
\underline{clear} $\rightarrow$ foggy   & \textbf{62.4}                  & 59.1                                  & 57.3                                 \\
\underline{clear} $\rightarrow$ rainy            & 54.6                           & 61.0                                  & \textbf{64.9}                        \\
\bottomrule
\end{tabular}
\caption{Comparison of different adaptation strategies for semantic segmentation under three directions of domain shift. The source domain is underlined. Incremental DA improves over Targeted DA, except for the case when Targeted DA underperforms the baseline. (Baseline = without DA)} \label{tab:incremental_vs_targeted}
\vspace{-1em}
\end{table}

\section{Experiments}
\thedataset allows studying the robustness of perception systems for driving under both discrete and continuous distributional shifts.
We first (\autoref{ssec:exp_discrete}) assess the impact of discrete domain shifts on model performance for multiple perception tasks available in our dataset and empirically demonstrate that observations from our simulation dataset transfer to real-world datasets. Moreover, we compare different discrete adaptation strategies and assess the calibration of uncertainty estimation methods under domain shifts. 
In \autoref{ssec:exp_continuous} we extend the analysis to continuous domain shifts and investigate properties of continuous domain adaptation methods~\cite{wang2020tent} against incremental adaptation and unsupervised domain adaptation~\cite{vu2019advent}.
%
%
Further experiments, implementation details, and ablations on the data collection choices are reported in the Appendix,
together with additional experiments on multitask learning.

\paragraph{Domain adaptation strategies.} \label{ssec:adaptation_strategies}
To analyze the impact of our dataset design choice, we examine the four domain adaptation strategies allowed by our dataset (\autoref{fig:adaptation_strategies}).
As \textit{Baseline}, we consider the model trained on the source domain only and directly tested on the other domains.
\textit{Targeted DA}~\cite{wang2018deep} is a traditional computer vision problem consisting of adapting from a labeled source domain to a specific unlabeled target domain.
We define \textit{Untargeted DA}~\cite{li2017deeper,segu2020batch} as adapting from a labeled source domain to a set of various unlabeled shifted domains. 
%
%
\textit{Incremental DA}~\cite{volpi2021continual}  consists in performing incremental steps of targeted DA from the source domain to the target domain passing through intermediate discretely-shifted domains.
\textit{Continuous TTA}~\cite{wang2020tent} aims at adapting frame by frame to a sequence presenting a continuously shifted domain from source to target domain.

\paragraph{Implementation details.} For the adaptation tasks, we focus on semantic segmentation and use ADVENT~\cite{vu2019advent} for the Targeted and Untargeted DA. 
The segmentation backbone is DRN-D-54~\cite{Yu2017}.
Incremental DA is performed as a series of Targeted DA steps, while for Continuous TTA we extend TENT~\cite{wang2020tent} to semantic segmentation and iteratively apply it on every incoming frame. Every model is trained in the clear-daytime domain and tested under different weather domains. While our dataset provides finer domain labels depending on the severity of the perturbation, we group different degrees of severity to match the environmental labels in BDD100K~\cite{bdd100k} in order to assess the compatibility of conclusions drawn from our dataset with real-world trends. 


\subsection{Discrete Shifts} \label{ssec:exp_discrete}
As outlined in \autoref{ssec:dataset_design}, our dataset provides incremental discrete shifts along natural environmental directions.
We investigate properties of discrete shifts on the multitude of supported tasks and report findings on domain adaptation and uncertainty estimation performance.
%

\paragraph{Impact of domain shift.} We find that many mainstream algorithms for different perception tasks suffer performance drops under domain shift (\autoref{tab:ood_performance}), where the severity increases with the distance from the source domain.
In particular, we train all models in the clear-daytime domain and test under different weather conditions, showing the overall negative impact of domain shift on all the vision tasks supported by our dataset. Nevertheless, in some specific cases a model may even perform better on a shifted domain, \eg{} instance segmentation on overcast. 
We leverage the incremental domain shifts provided in our dataset to investigate in \autoref{tab:incremental_vs_targeted} different discrete adaptation strategies for semantic segmentation, \ie{} Incremental DA and Targeted DA. We find that incrementally adapting from source to target domain improves the generalization to the target domain compared to direct Targeted DA. 
%
However, clear~$\xrightarrow{}$~foggy represents a challenging scenario for which both the adaptation strategies worsen the baseline performance.
%

\paragraph{Real-world compatibility.} To establish a reliable benchmark we must first confirm that trends witnessed in our simulation dataset are compatible with real-world observations.
We use BDD100K~\cite{bdd100k} for  comparison because it features the largest subset of our tasks available in a real-world dataset with discrete domain labels.
We study the domain shift effect on two fundamental perception tasks, \ie{} 2D object detection and semantic segmentation, and show compatible trends for different methods trained on \thedataset{} and BDD100K (\autoref{fig:trend}). 
We evaluate the one-stage method YOLO v3~\cite{redmon2018yolov3}, as well as the two-stage methods Faster R-CNN~\cite{ren2015faster} and Cascade R-CNN~\cite{cai2018cascade} for object detection. For semantic segmentation, we consider three different methods, FCN~\cite{long2015fully}, DRN-D~\cite{yu2017dilated}, and DeepLab v3+~\cite{chen2017deeplab}. 
%
Our experiments suggest that the performance of different methods for semantic segmentation and object detection degrades under different domain shifts. Moreover, we find that the ranking of methods and the relative degradation trend is compatible between SHIFT and the real-world dataset BDD100K, confirming the usefulness of SHIFT and its consistency with the real world.
%

\begin{table}[t]
\centering
\footnotesize
\setlength{\tabcolsep}{8pt}
\setlength{\tabcolsep}{4pt}
\begin{tabular}{llcccccccc}
 & \textbf{Method}  &
  \parbox[t]{8mm}{\rotatebox[origin=l]{90}{\shortstack[l]{\underline{clear-}\\\underline{daytime}}}} &
  \parbox[t]{2mm}{\rotatebox[origin=l]{90}{cloudy}} &
  \parbox[t]{2mm}{\rotatebox[origin=l]{90}{overcast}} &
  \parbox[t]{2mm}{\rotatebox[origin=l]{90}{foggy}} &
  \parbox[t]{2mm}{\rotatebox[origin=l]{90}{rainy}} &
  \parbox[t]{2mm}{\rotatebox[origin=l]{90}{dawn/\\dusk}} &
  \parbox[t]{2mm}{\rotatebox[origin=l]{90}{night}} & 
  \parbox[t]{5mm}{\rotatebox[origin=l]{90}{\textbf{\shortstack[l]{OOD\\avg.}}}} \\ \midrule
\parbox[t]{2mm}{\multirow{3}{*}{\rotatebox[origin=c]{90}{\textbf{SHIFT}}}}  & Softmax  & 3.3 & 32.6 & 14.2 & 48.8 & 64.3 & 43.7 & 64.7 & 45.2\\
& MCDO & 1.2 & 13.1 & 7.6 & 20.8 & 10.0 & 27.2 & 39.6 & 19.7 \\
    & Ensemble & 1.4 & 12.3 & 7.5 & 23.4 & 8.9 & 18.7 & 36.9 & 18.0 \\\midrule
\parbox[t]{2mm}{\multirow{3}{*}{\rotatebox[origin=c]{90}{\textbf{BDD}}}} & Softmax & 9.6 & 23.2 & 9.9 & 9.7 & 7.7 & 10.6 & 48.6 & 18.4 \\
& MCDO &  12.3 & 22.0 & 7.8 & 13.0 & 11.4 & 13.1 & 41.4 & 18.1 \\
& Ensemble & 12.6 & 18.8 & 9.2 & 11.7 & 11.8 & 13.9 & 39.8 & 17.5\\
                      \bottomrule
\end{tabular}
\caption{Calibration (ECE, \%) of uncertainty estimation methods under distributional shift for semantic segmentation. The lower, the better. Source domain is clear-daytime. We find that calibration worsens far from the source, both for \thedataset and BDD100K.} \label{tab:segmentation_uncertainty} 
\vspace{-3mm}
\end{table}

\paragraph{Uncertainty estimation.} Autonomous driving systems must deal with life-threatening failure cases. To this end, uncertainty estimation represents a powerful tool to assess the reliability of a model's predictions. 
%
%
Following~\cite{guo2017calibration}, we evaluate the Expected Calibration Error (ECE) to assess the calibration, \ie{} correlation with model error, of uncertainty estimation methods under domain shift.
In particular, we evaluate the Softmax Entropy baseline and traditional Bayesian techniques such as Monte-Carlo Dropout (MCDO)~\cite{gal2016dropout} and Deep Ensembles~\cite{lakshminarayanan2017simple}.
We observe that such uncertainty estimation methods are not well calibrated under domain shift, and that calibration worsens under incremental shifts on both \thedataset and BDD100K (\autoref{tab:segmentation_uncertainty}).
%
%
While some domains are more challenging in \thedataset than in BDD100K, the overall degradation of calibration observed on \thedataset is confirmed on BDD100K and the ranking of methods is preserved, further highlighting that conclusions drawn from our dataset transfer to the real world.
%

%
%

%
We hope that our dataset will help researchers providing solutions to the potentially life-threatening shortcomings of current DA and uncertainty estimation techniques.

\begin{figure}[t]
    \centering
    \includegraphics[width=1\linewidth]{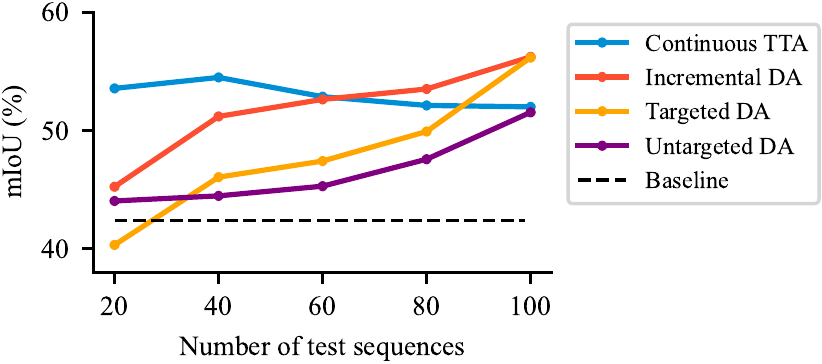}
  \caption{Comparison of different adaptation strategies for semantic segmentation on daytime~$\xrightarrow{}$~night shifts at varying amounts of available sequences. TTA is the most effective under limited amounts of data. When enough data becomes available, Incremental DA outperforms all other alternatives.}
    \label{fig:compare_adaptation_strategies}
\end{figure}

\begin{figure}[t]
    \centering
    \includegraphics[width=1.0\linewidth]{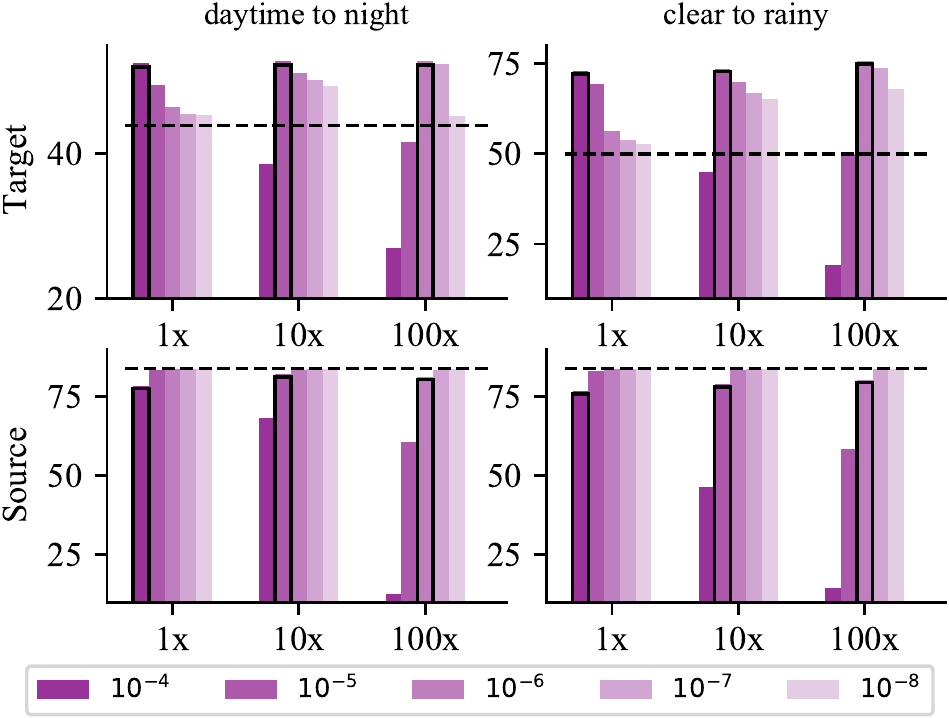}
    \caption{Performance on the target domain of TTA for different sequence lengths. Best learning rate on target domain is highlighted by black boxes. Both source and target performance are highly sensitive to the learning rates. Dashed lines = before TTA.}
    \label{fig:tta_histograms}
    \vspace{-1em}
\end{figure}

\begin{figure*}[t]
    \centering
    \vspace{-1em}
    \includegraphics[width=1\linewidth]{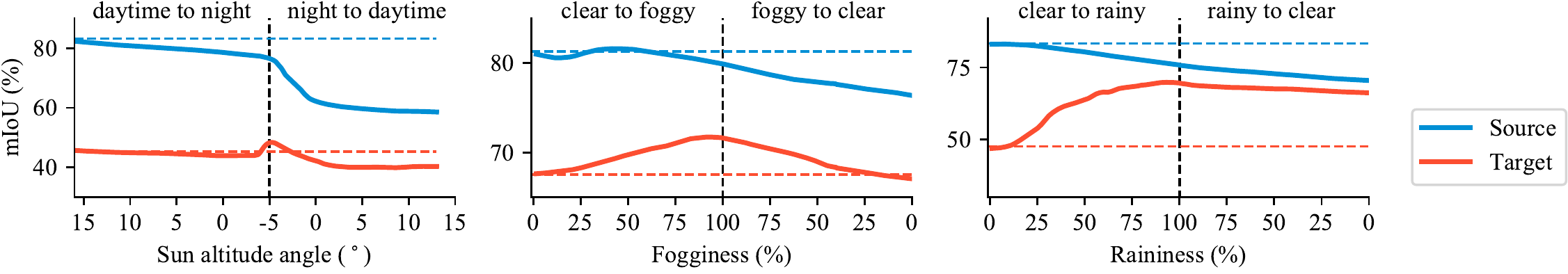} \\
    \includegraphics[width=1\linewidth]{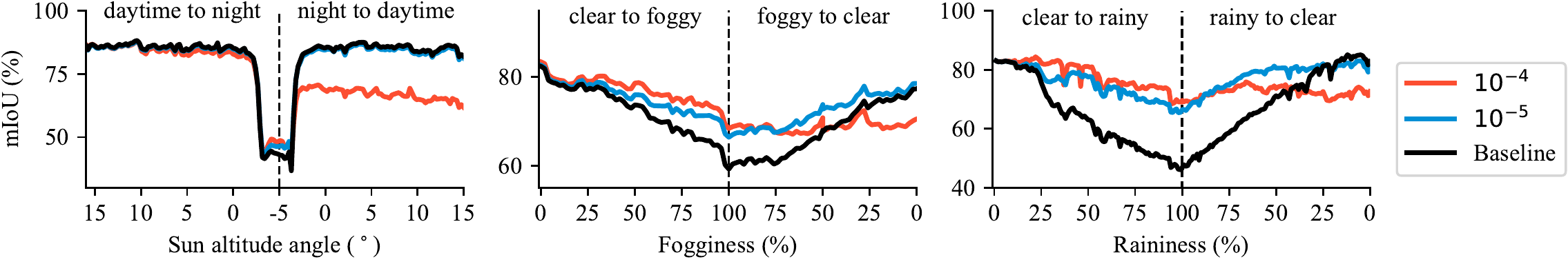}
  \caption{Performance of TTA for semantic segmentation under three types of domain shift: daytime $\xrightarrow{}$ night, clear $\xrightarrow{}$ foggy, clear $\xrightarrow{}$ rainy. Each point corresponds to the performance of the model on the source (top-blue) / target (top-red) / current (bottom) domain finetuned up to that level of domain shift in the sequence. Horizontal lines in the bottom figure represent the original performance on source (blue) and target domain (red). After reaching the target domain, every sequence loops back to the original source domain. Catastrophic forgetting can be observed by the drop in source performance during TTA.}
    \label{fig:tta_loop_back}
\end{figure*}

\subsection{Continuous Shifts} \label{ssec:exp_continuous}
A key feature of \thedataset{} is that of providing a set with continuous intra-sequence domain shifts, allowing to compare different adaptation strategies under continuous shifts and provide an in-depth analysis on TTA and its properties.

\paragraph{Continual domain adaptation.} \autoref{fig:compare_adaptation_strategies} compares four different adaptation strategies for semantic segmentation on an increasing number of sequences.
Given a model pretrained on the source domain, \ie{} clear-daytime, and the set of continuously shifting sequences from one domain to another, \ie{} clear-daytime~$\xrightarrow{}$~night, we train the TTA algorithm on each frame of the incoming data stream. TTA is thus performed independently on each sequence. Final performance is averaged over all the sequences.
For the other adaptation strategies, we divide the length of the sequence in 20 bins, consider each bin as a separate domain, and group corresponding bins from all the provided sequences. For Targeted DA, we thus adapt directly to the last bin, corresponding to the night domain. Untargeted DA is instead applied on all the bins but the source one. Incremental DA is performed by incrementally adapting from one bin to the consecutive one until the end of the sequence is reached. 
In particular, we plot the average mIoU against the number of training sequences (\autoref{fig:compare_adaptation_strategies}).
We find that TTA is extremely efficient under small target data availability compared to all other alternatives, and that Incremental DA is consistently more effective than both Targeted and Untargeted DA.

\paragraph{Test-time adaptation.} As intra-sequence continuous shifts 
represent one of the main contributions of \thedataset, we further focus on TTA by using TENT~\cite{wang2020tent} and evaluate the effect of the speed at which domain shift happens within a sequence on TTA performance (\autoref{fig:tta_histograms}). This is made possible by the sets of sequences of different lengths (1x, 10x, 100x the basic sequence length). 

Given a source and a target domain, \eg{} daytime and night, each sequence starts from the source domain and reaches the target domain at mid-sequence length; then, it loops back to the original domain. 
We first observe that, depending on the domain shift speed, the learning rate can highly affect the outcome of the TTA (\autoref{fig:tta_histograms}). Slower (faster) shifts will require lower (higher) learning rates. 
Moreover, after reaching the target domain at mid-sequence, the performance on the target domain has improved compared to its original value, while that on the source domain has dropped.  
According to \autoref{fig:tta_histograms} (1x), we find that the optimal learning rate in terms of adaptation to the target domain leads to the largest performance drop on the original source (\autoref{fig:tta_loop_back}, top). This problem, known as catastrophic forgetting~\cite{kirkpatrick2017overcoming} in the continual learning literature, has already been observed for class- and task-incremental learning.

To further investigate this issue, we loop back to the original domain after adapting to the target and find that, while the performance on the current target domains largely improves over the baseline (\autoref{fig:tta_loop_back}, bottom), the original source domain accuracy cannot be recovered (\autoref{fig:tta_loop_back}, top).
While TTA has shown to be extremely effective to adapt on the go, a model adapted with TTA cannot be safely deployed on the original source domain.
Showing that catastrophic forgetting also affects test-time adaptation further demonstrates the importance of providing continuously shifted sequences in driving datasets, and we hope that future research will attempt to mitigate this problem. 

%


\section{Conclusion}
We introduce \thedataset{}, a multi-task driving dataset featuring the most important perception tasks under discrete and continuous domain shifts.
Thanks to our dataset design, we demonstrate several new findings on different adaptation strategies and uncertainty estimation methods.
Although simulation environments are still far from being a perfect representation of the real world, they allow inexpensive data collection and annotation.
Moreover, we empirically demonstrate that conclusions drawn from our dataset hold in real-world datasets.
To the best of our knowledge, \thedataset{} is the largest synthetic dataset for autonomous driving, providing the most inclusive set of annotations and conditions.
We hope that providing the first dataset with realistic continuous domain shifts will contribute to shaping the data collection paradigm for real-world driving datasets and promote advances in test-time learning and adaptation.

\clearpage
{\small
\bibliographystyle{ieee_fullname}
\bibliography{egbib}
}

\clearpage

\section*{Appendix}
We provide additional details on our dataset in \autoref{sec:dataset_details}.
In particular, we report the sensor layout (\autoref{ssec:dataset_sensors}), annotation details (\autoref{ssec:dataset_annotations}), extensive information on dataset generation (\autoref{ssec:dataset_generation}) and dataset statistics (\autoref{ssec:dataset_statistics}).

Moreover, we conduct additional experiments in \autoref{sec:suppl_exp}. 
We provide baselines on multitask learning under continual domain shift (\autoref{sec:suppl_exp_multitask}), and conduct ablation studies on joint training with real-world data (\autoref{sec:suppl_exp_joint_training}) and the optimal dataset size for each task (\autoref{sec:suppl_exp_dataset_size}). Further, we propose a qualitative comparison between properties of \thedataset{} and the VIPER dataset~\cite{Richter_2017} (\autoref{sec:viper}), and ablate on the model failures on the rainy and foggy domains (\autoref{sec:error}).

Implementation details for each experiment conducted in this work are reported in \autoref{sec:implementation_details} for full reproducibility.

\def\thesection{\Alph{section}}
\setcounter{section}{0}

\begin{figure}[b]
    \centering
    \includegraphics[width=\linewidth]{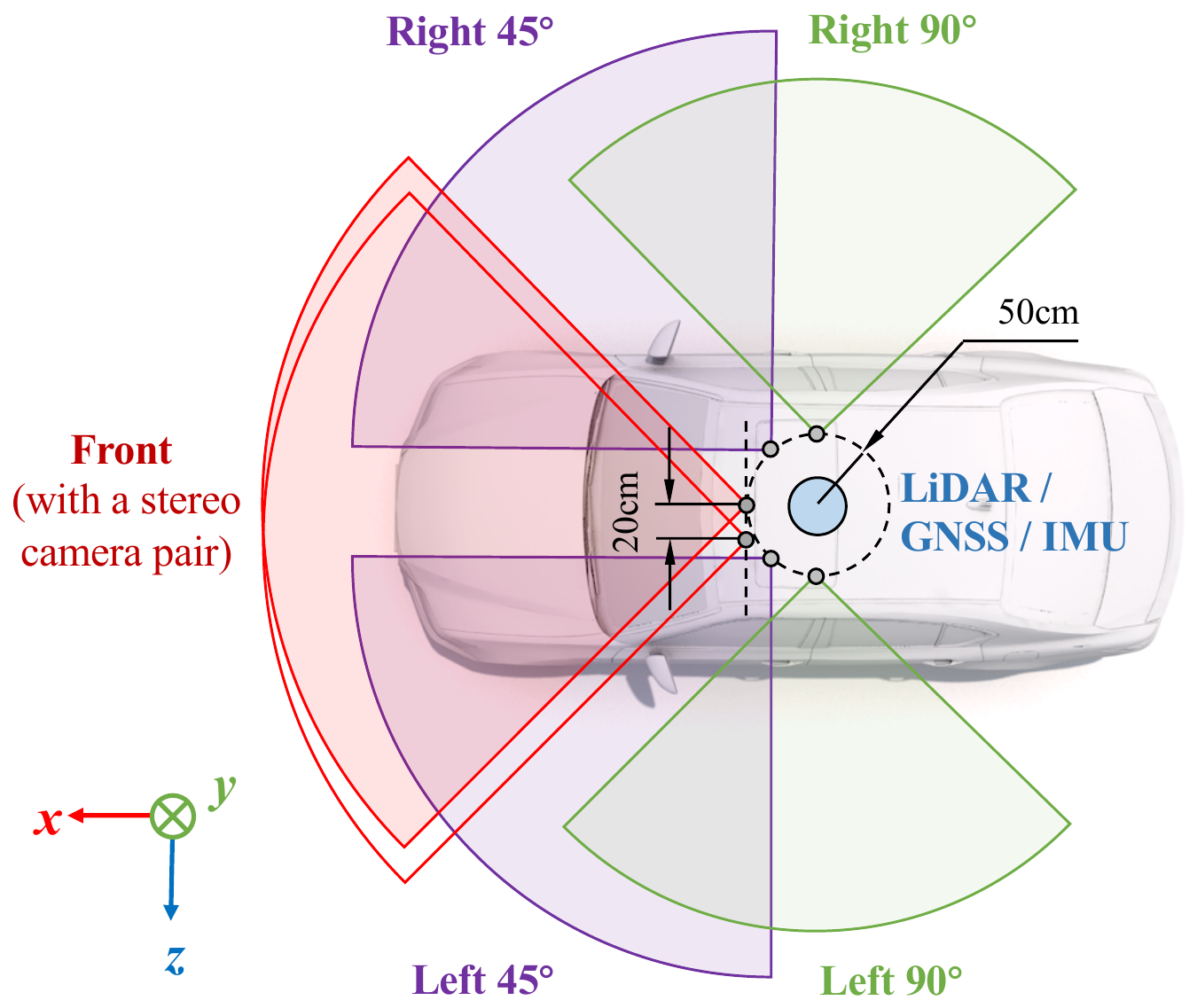}
    \caption{The vehicle system and the sensor layout. Except for stereo cameras, all the cameras are located on a circle centered at the vehicle reference system's origin (blue dot). LiDAR and motion sensors are located at the origin. Axes directions of the vehicle system are shown at the bottom left corner. \textit{Best viewed in color.}}
    \label{fig:sensors}
\end{figure}

\section{Dataset Details} \label{sec:dataset_details}
The detailed user guide and additional information can be found at \url{https://www.vis.xyz/shift}.

\subsection{Reference systems and sensor layout} \label{ssec:dataset_sensors}
The dataset has three levels of reference systems: \textit{world}, \textit{vehicle}, and \textit{camera}. The world system represents the absolute position of objects. The vehicle system is used for storing all 3D annotations. The camera systems are the reference systems used for each individual camera.

\autoref{tab:sensors} summarizes the supported sensors. We set up the vehicle system following KITTI's convention and the right-hand rule.
Specifically, the origin is located at the center of the ego-vehicle (marked as the blue dot in \autoref{fig:sensors}).
Its \textit{x}, \textit{y}, and \textit{z} axes point in the right, down, and front directions, respectively (\autoref{fig:sensors}, bottom left). 
All the sensors are located on a circle centered at the vehicle reference system's origin, except for the stereo cameras that are placed on the left to the front camera, with a horizontal displacement of $20 \mathrm{cm}$. 
All the cameras have a field-of-view (FoV) of $90\degree$.
The 128-channel LiDAR sensor has a vertical FoV range of $[-10\degree, +10\degree]$ and a scan rate of 1.12M points per second. 

\begin{table}[t]
    \centering
    \footnotesize
    \begin{tabular}{ll p{0.48\linewidth}}
    \toprule
      \textbf{Sensor}  &  \textbf{Data type} & \textbf{Position} \\
         \midrule
        RGB camera & 24-bit RGB & 5 $\times$ RGB cameras (front, left / right $45\degree$, left / right $90\degree$). \\
        Stereo camera & 24-bit RGB & Additional RGB camera offsetting $20\mathrm{cm}$ toward left from the center. \\
        Depth camera & 24-bit Gray & Same as front view RGB camera. \\
        Optical flow & 32-bit UV & Same as front view RGB camera. \\
        GNSS / IMU & Vector & Center of vehicle. \\
    \bottomrule
    \end{tabular}
    \caption{The data type and position settings of the sensors.}
    \label{tab:sensors}
\end{table}

Annotations stored in the vehicle system can be easily converted into the camera systems. Here, the front cameras and LiDAR sensor have camera systems identical to the vehicle system, so no conversion is needed for them. For other cameras, a vehicle-to-camera matrix (\ie{} intrinsic and extrinsic parameters) is provided to transform the annotations so that they fit each camera. 


\subsection{Annotation details} \label{ssec:dataset_annotations}
We present detailed specifications for the annotation set provided in \thedataset{}.

\paragraph{Object detection} is a fundamental localization task for scene understanding and a basis for numerous downstream driving tasks, including multiple object tracking (MOT) and object re-identification (ReID). 
We provide 2D/3D bounding box annotations and object identities for six categories of traffic participants, \ie{} car, truck, bus, bicycle, motorcycle, and pedestrian, together with the visibility attributes `occluded' and `truncated'.
Moreover, for each box, we provide fine-grained object classes (\eg{} vehicle model type).

%
While previous datasets only provide 7 DoF (\ie{} only yaw angle) 3D boxes~\cite{geiger2013vision,caesar2020nuscenes,sun2020scalability}, we provide 9 DoF annotations and use the Euler angle system (\textit{i.e.}, yaw, roll, pitch) to represent the orientation for bounding boxes in 3D space.


\paragraph{Image segmentation} is a fundamental pixel-level perception task. 
For each frame, we provide panoptic (\ie{} instance and semantic) segmentation labels on the 23 classes of the Cityscapes~\cite{cordts2016cityscapes} annotation scheme. 
Together with 2D bounding boxes, segmentation labels can be used in multi-object tracking and segmentation (MOTS) and multi-object panoptic tracking (MOTP) tasks.

\begin{table*}[t]
\footnotesize
\setlength{\tabcolsep}{3pt}
\begin{tabular}{@{}lllll@{}}
\toprule
\textbf{Category} $\mathcal{H}_i$ &
  \textbf{Candidate dom.} $h_i^{(j)}$ &
  \textbf{BDD100K eq.} &
  \textbf{Environmental parameters } &
  \textbf{Degrees of shift} \\ \midrule
\multirow{6}{*}{Time of day} &
  {noon} &
  \multirow{2}{*}[1pt]{$\left.\rule{0cm}{0.3cm}\right\}$ daytime} &
  Sun altitude angle = \{90, 75, 60, 45, 30\} &
  \multirow{6}{*}{altitude angle $\in [-5, 90]$} \\
 & morning / afternoon &                                & Sun altitude angle = \{15, 10, 5\}                                    &  \\ 
 & dawn / dusk         & \multirow{2}{*}[1pt]{$\left.\rule{0cm}{0.3cm}\right\}$ dawn / dusk} & Sun altitude angle = \{4, 3, 2\}                                      &  \\
 & sunrise / sunset    &                                & Sun altitude angle = \{1, 0, -1\}                                     &  \\ 
 & night               & \multirow{2}{*}[1pt]{$\left.\rule{0cm}{0.3cm}\right\}$ night}         & Sun altitude angle = \{-2, -3\}                                       &  \\
 & {dark night}          &                                & Sun altitude angle = \{-4, -5\}                                       &  \\ \midrule
\multirow{10}{*}{Weather} &
  {clear} &
  \multirow{2}{*}[1pt]{$\left.\rule{0cm}{0.3cm}\right\}$ clear}  &
  cloudiness = \{0, 5\} &
  \multirow{4}{*}{cloudiness $\in [0, 100]$} \\
 & slight cloudy       & 
& cloudiness = \{10, 15\}        
&  \\
 & partly cloudy              &     partly cloudy                           & cloudiness = \{25, 50, 70\}                                               &  \\
 & {overcast}            & overcast                       & cloudiness = 100                                                      &  \\ \cmidrule(rr){2-5} 
 &
  {small rain} &
  \multirow{3}{*}{rainy} &
  cloudiness = 70; precipitation = 20; deposit = 60; fog den. = 3 &
  \multirow{3}{*}{precipitation $\in [0, 100]$} \\
 & mid rain            &                                & cloudiness = 80; precipitation = 50; deposit = 80; fog den. = 3    &  \\
 & {heavy rain}          &                                & cloudiness = 100; precipitation = 100; deposit = 100; fog den. = 7 &  \\ \cmidrule(l){2-5} 
 &
  {small fog} &
  \multirow{2}{*}{foggy} &
  cloudiness = 60; fog density = 30; fog distance = 15 &
  \multirow{2}{*}{fog density $\in [0, 100]$} \\
 & {heavy fog}           &                                & cloudiness = 80; fog density = 90; fog distance = 20                  &  \\ \midrule
\multirow{3}{*}{Vehicle density} &
  sparse &
  - &
  num of vehicle = 50 &
  \multirow{3}{*}{\begin{tabular}[c]{@{}l@{}}vehicle per map,\\vehicle per frame\end{tabular}} \\
 & moderate            & -                              & num of vehicle = 100                                                  &  \\
 & crowded             & -                              & num of vehicle = 250                                                  &  \\ \midrule
\multirow{3}{*}{Pedestrian density} &
  sparse &
  - &
  num of pedestrians = 100 &
  \multirow{3}{*}{\begin{tabular}[c]{@{}l@{}}pedestrian per map,\\ pedestrian per frame\end{tabular}} \\
 & moderate            & -                              & num of pedestrians = 200                                              &  \\
 & crowded             & -                              & num of pedestrians = 400                                              &  \\ \bottomrule
\end{tabular}
\caption{Definitions of the domain category and candidate domains, used for discrete domain shifts. Each category has a group of candidate domains.  For each candidate domain, we show its equivalent domain label in BDD100K and the environmental parameters for simulation. }
\label{tab:parameters}
\end{table*}

\paragraph{Depth estimation} is an essential step to extend the 2D perception tasks into the 3D setting. We provide the depth labels aligned with the front-view RGB camera to enable image- and video-based monocular and stereo depth estimation. Depth resolution is $1\mathrm{mm}$.

\paragraph{Optical flow estimation} is an essential task for driving algorithms involving motion. However, existing large-scale datasets typically do not provide optical flow annotations due to the high labeling cost. Representing the relative motion between each pixel in a pair of images, optical flow can be instrumental in object tracking and ego-motion tasks. 
We provide the optical flow labels in the UV map format, also used in KITTI~\cite{geiger2013vision}.

\subsection{Data generation pipeline} \label{ssec:dataset_generation}

We introduce the pipeline that used to generate the discrete and continuous domain shifts.

\paragraph{Disrete shift.}
As discussed in \autoref{ssec:dataset_design}, we set up an efficient sampling pipeline that can cover a diverse combination of conditions. To determine the environmental parameters of each sequence, we use a technique similar to random search. In \autoref{tab:parameters}, we define 4 categories of domain shifts, \textit{e.g.}, time of day, weather, vehicle density, and pedestrian density. For the $i$-th category ($1\leq i \leq 4$), we define a 
set of \textit{candidate} domains, $\mathcal{H}_{i} = \{h_i^{(1)}, \cdots , h_i^{(n_i)} \}$, where each candidate $h_i^{(j)}$ corresponds to a certain group of environmental parameters, defined in the \autoref{tab:parameters}. Note that the parameter can be a fixed value or a set of values. For the set of values, we again uniformly sample one value out of the set. 

Our sampling method for the discrete domain shifts can be summarized as following. A sequence is generated with a fixed parameter vector $\bm{\theta} = \theta_1 \cup \cdots \cup \theta_m$, where each $\theta_i$ is sampled uniformly across all candidates in the $i$-th category, \textit{i.e.},
\begin{align}
    h_i^{(j)} & \sim \mathrm{Uniform}(\mathcal{H}_{i}), \quad \forall i = \{1,2,3,4\} \\
    \theta_i & \sim h_i^{(j)}
\end{align}
This pipeline guarantees the uniform marginal distribution of candidates conditioned on any category. Using this pipeline, we can easily add data without breaking the distribution of domains. Moreover, any subset of sequences of \thedataset{} has the same distribution, allowing a fair experiment on the impact of data amount.


\begin{table*}[t]
\footnotesize
\centering
\setlength{\tabcolsep}{14pt}
\begin{tabular}{@{}llll@{}}
\toprule
\multirow{2}{*}{\textbf{Continuous shift type}} & \multicolumn{3}{c}{\textbf{Environmental parameters}}                          \\ \cmidrule(l){2-4} 
                                                & \textbf{Beginning state} ($t=0$)                   & \textbf{Intermediate state} ($t=0.2$)  & \textbf{End state} ($t=1$)                     \\ \midrule
Time of day                                     & Sun altitude angle = 90 & -            & Sun altitude angle = -5 \\ \hline
Cloudiness                                      & cloudiness = 0          &  -            & cloudiness = 100        \\ \hline
Raininess &
  \begin{tabular}[c]{@{}l@{}}cloudiness = 0, precipitation = 0, \\ deposit = 0, fog density = 0\end{tabular} &
  \begin{tabular}[c]{@{}l@{}}cloudiness = 80, precipitation = 50,\\ deposit = 80, fog density = 3\end{tabular} &
  \begin{tabular}[c]{@{}l@{}}cloudiness = 100, precipitation = 100,\\ deposit = 100, fog density = 7\end{tabular} \\ \hline
Fogginess &
  \begin{tabular}[c]{@{}l@{}}cloudiness = 0, fog density = 0,\\ fog distance = 0\end{tabular} &
  \begin{tabular}[c]{@{}l@{}}cloudiness = 60, fog density = 30,\\ fog distance = 15\end{tabular} &
  \begin{tabular}[c]{@{}l@{}}cloudiness = 80, fog density = 90,\\ fog distance = 20\end{tabular} \\ \bottomrule
\end{tabular}
\caption{Definitions of parameters used for continuous domain shifts. The parameters are updated for every frames during driving. The value of parameters are determined by linear interpolation between the state of beginning, intermediate (if applicable) and end. }
\label{tab:parameters_continuous}
\end{table*}
\paragraph{Continuous shifts.} For sequences with continuous domain shift, the change of parameters happens on one specific domain category $c$, while others are kept unchanged, \ie{} the frame at time $t$ is generated with the parameter vector
\begin{equation}
    \bm{\theta}(t) = \theta_1 \cup \cdots \cup f_c(t) \cup \cdots \cup \theta_4 \; ,
\end{equation}
where $f_c(t)$ is obtained by linear interpolation of the states listed in the `Environmental parameters' column in  \autoref{tab:parameters}. Specifically, $f_c(t)$ is obtained by interpolating the points
\begin{equation}
   (t, \theta) = [ \left(0, \theta_{c, \mathrm{begin}}\right),  \left(0.2, \theta_{c, \mathrm{intermediate}}\right), \left(1, \theta_{c, \mathrm{end}}\right)] \; ,
\end{equation}
where $t \in [0, 1]$ represents the degree of continuous shift from the minimum to the maximum parameter allowed for a given domain category.

Our dataset provides 300, 120, 30 continuous shift sequences in 1x, 10x, 100x length respectively, where the base length (1x) is 200 frames. Since continuous domain shift is generated by interpolating between the state of the initial frame and the state of the final frame, the domain shift speed is inversely proportional to the sequence length in frames.
Furthermore, we provide an additional set of 150 sequences of base length presenting domain shifts simultaneously happening along multiple domain shift directions within the same sequence.

\paragraph{Domain labeling details.}
The degree of shift for each domain category is quantified by a numerical value called \textit{severity}. For weather conditions, we use percentage values to indicate the degree of severity, where $0\%$ corresponds to clear weather conditions and $100\%$ represents the most extreme condition allowed by the CARLA simulator for a given weather direction, \eg{} cloudiness, precipitation, fog density, or fog distance. We describe the time of day using the Sun's altitude angle to disentangle the lighting condition with the sunrise/sunset time. For the object densities, we use the number of objects per frame as the severity (\autoref{tab:parameters}).

\subsection{Dataset statistics} \label{ssec:dataset_statistics}
\begin{figure}[t]
    \centering
    \includegraphics[width=0.49\linewidth]{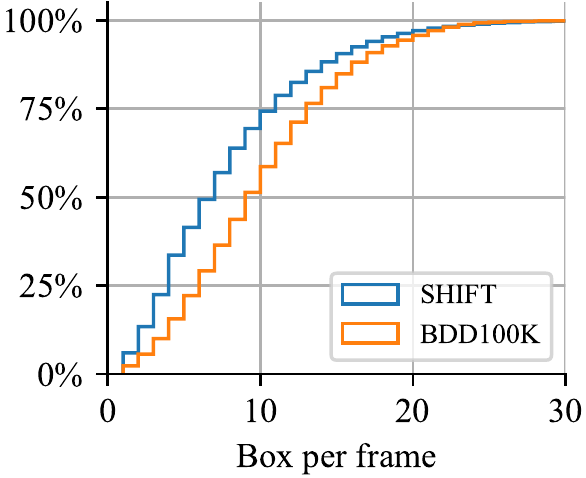} \hfill
    \includegraphics[width=0.48\linewidth]{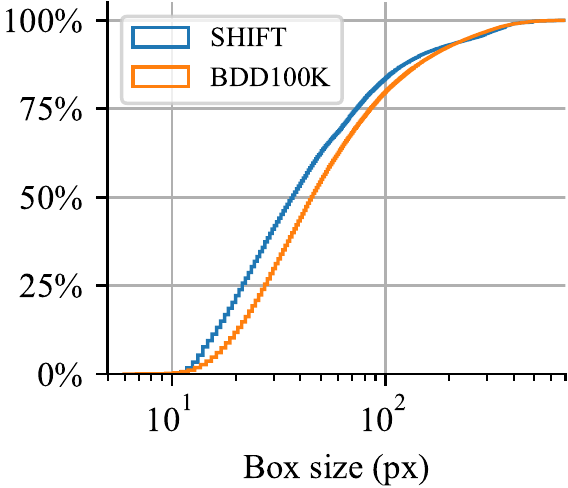}
    \caption{Cumulative distributions of the bounding box per frame (left) and  bounding box size measured in $\sqrt{wh}$ (right). We only count the objects in the front camera view.  \thedataset{} covers various object densities and a wide range of object scale.}
    \label{fig:bounding_box}
\end{figure}

\thedataset{} is diverse in bounding box scale. \autoref{fig:bounding_box} (left) plots the object density measured by boxes per frame and shows coverage from 0 to 30 boxes/frame for \thedataset{}. We compare the distribution with the BDD100K's MOT set. Due to the sparsity of the vehicle/pedestrians density domains, our dataset has on average a higher density of frames counting less bounding boxes than BDD100K, but the crowded frames ($\geq20$ boxes/frame) show similar trends. Moreover, \autoref{fig:bounding_box} (right) shows the distribution of bounding box sizes, defined as $\sqrt{wh}$ where $w$ and $h$ are the width and height of a box. \thedataset{} covers diverse box sizes ranging from 10 to 650 pixels. We also observe that our dataset has 41.2\% bounding boxes smaller than 15 pixels while BDD100K has 30.9\%, showing that our dataset provides challenging conditions for small object detection and tracking.

\begin{table*}[t]
\setlength{\tabcolsep}{3.3pt}
\center
\footnotesize
\begin{tabular}{@{}lllcccccccc|ccc@{}}
\toprule
\multirow{2}{*}{\textbf{Task}}                        & \multirow{2}{*}{\textbf{Train}} & \multirow{2}{*}{\textbf{Metric}}                                                   & \multicolumn{1}{c}{\textbf{Source}} & \multicolumn{6}{c}{\textbf{OOD}}                                                                           & \multicolumn{1}{c}{\multirow{2}{*}{\textbf{OOD Avg.}}} & \multicolumn{1}{c}{\multirow{2}{*}{$\Delta_{\text{Source}}$}} & \multicolumn{1}{c}{\multirow{2}{*}{$\Delta_{\text{OOD}}$}} & \multicolumn{1}{c}{\multirow{2}{*}{$\Delta_{\text{S} \rightarrow \text{O}}$}} \\  \cmidrule(lr){4-4} \cmidrule{5-10} 
                                           &                        &                           & {clear-daytime}                      & {cloudy} & {overcast} & {foggy} & {rain} & {dawn/dusk} & {night} & \multicolumn{1}{l}{}                         & \multicolumn{1}{l}{}                          & \multicolumn{1}{l}{}                       & \multicolumn{1}{l}{}                       \\ \midrule
\multirow{8}{*}{\begin{tabular}[c]{@{}l@{}}Semantic \\ segmentation (S)\end{tabular}}   & S                      & \multirow{8}{*}{mIoU (\%) $\uparrow$}                                                & 69.1                                & 40.6            & 40.6              & 21.5           & 19.6          & 18.1               & 8.9            & 24.9                                         & -                                             & -                                          & -64.0\%                                    \\
                                                                                        & S + D                  &                                                                           & \textbf{75.2}                       & 53.8            & 52.6              & 24.3           & 26.6          & 24.0               & 9.9            & 31.9                                         & \textbf{8.9\%}                                & 28.1\%                                     & -57.6\%                                    \\
                                                                                        & S + F                  &                                                                           & 69.4                                & 51.8            & 54.7              & 26.4           & 22.4          & 22.7               & 9.8            & 31.3                                         & 0.4\%                                         & 25.8\%                                     & -54.9\%                                    \\
                                                                                        & S + D + F              &                                                                           & 71.8                                & 50.0            & 51.9              & 23.5           & 24.0          & 22.1               & 9.5            & 30.2                                         & 3.8\%                                         & 21.2\%                                     & -58.0\%                                    \\
                                                                                        & S + I                  &                                                                           & 74.8                                & \textbf{63.9}   & \textbf{68.1}     & \textbf{41.0}  & \textbf{36.8} & \textbf{37.3}      & \textbf{23.6}  & \textbf{45.1}                                & 8.2\%                                         & \textbf{81.3\%}                            & \textbf{-39.7\%}                           \\
                                                                                        & S + D + I              &                                                                           & 75.0                                & 62.4            & 65.1              & 37.4           & 35.4          & 35.3               & 20.5           & 42.7                                         & 8.6\%                                         & 71.6\%                                     & -43.1\%                                    \\
                                                                                        & S + F + I              &                                                                           & 72.5                                & 58.3            & 59.6              & 35.8           & 27.2          & 28.8               & 14.4           & 37.3                                         & 4.9\%                                         & 50.1\%                                     & -48.5\%                                    \\
                                                                                        & S + D + F + I          &                                                                           & 74.7                                & 60.6            & 59.6              & 37.1           & 32.7          & 33.1               & 19.3           & 40.4                                         & 8.1\%                                         & 62.3\%                                     & -45.9\%                                    \\ \midrule
\multirow{8}{*}{\begin{tabular}[c]{@{}l@{}}Depth\\ estimation (D)\end{tabular}}                                                   & D                      & \multirow{8}{*}{SILog $\downarrow$}                                                    & 17.8                                & 28.3            & 23.1              & 81.9           & 46.3          & 54.6               & 63.2           & 49.6                                         & -                                             & -                                          & -64.1\%                                    \\
                                                                                        & D + S                  &                                                                           & 16.9                                & 25.2            & 22.4              & 65.7           & 43.0          & 49.4               & 57.6           & 43.9                                         & 5.6\%                                         & 13.0\%                                     & -61.6\%                                    \\
                                                                                        & D + F                  &                                                                           & 19.3                                & 25.3            & 20.4              & 66.6           & 45.3          & 50.3               & 54.4           & 43.7                                         & -7.8\%                                        & 13.4\%                                     & -55.8\%                                    \\
                                                                                        & D + S + F              &                                                                           & 19.6                                & 26.9            & 24.7              & 67.8           & 45.1          & 52.1               & 56.6           & 45.5                                         & -9.2\%                                        & 8.9\%                                      & -56.9\%                                    \\
                                                                                        & D + I                  &                                                                           & 17.3                                & 21.0            & 16.8              & 66.4           & 35.1          & 42.4               & 48.4           & 38.3                                         & 2.9\%                                         & 29.3\%                                     & -54.8\%                                    \\
                                                                                        & D + S + I              &                                                                           & \textbf{16.0}                       & \textbf{19.5}   & \textbf{15.4}     & 61.1           & \textbf{31.2} & \textbf{38.5}      & \textbf{42.7}  & \textbf{34.7}                                & \textbf{11.0\%}                               & \textbf{42.8\%}                            & -53.8\%                                    \\
                                                                                        & D + F + I              &                                                                           & 17.8                                & 21.4            & 17.9              & \textbf{47.9}  & 36.4          & 39.9               & 46.3           & 35.0                                         & 0.1\%                                         & 41.8\%                                     & \textbf{-49.1\%}                           \\
                                                                                        & D + S + F + I          &                                                                           & 17.6                                & 21.9            & 18.3              & 53.2           & 37.1          & 42.7               & 50.5           & 37.3                                         & 1.0\%                                         & 33.0\%                                     & -52.7\%                                    \\ \midrule
\multirow{8}{*}{\begin{tabular}[c]{@{}l@{}}Optical flow \\ estimation (F)\end{tabular}} & F                      & \multirow{8}{*}{EPE (px) $\downarrow$}                                                 & \textbf{6.0}                        & \textbf{6.7}    & \textbf{6.4}      & \textbf{9.0}   & \textbf{9.7}  & \textbf{9.1}       & \textbf{11.0}  & \textbf{8.6}                                 & \textbf{-}                                    & \textbf{-}                                 & -30.8\%                                    \\
                                                                                        & F + S                  &                                                                           & 7.8                                 & 8.3             & 8.5               & 10.4           & 12.0          & 10.9               & 12.6           & 10.4                                         & -23.1\%                                       & -17.4\%                                    & -25.7\%                                    \\
                                                                                        & F + D                  &                                                                           & 6.0                                 & 6.9             & 6.4               & 9.4            & 10.4          & 9.6                & 11.8           & 9.1                                          & -0.2\%                                        & -5.0\%                                     & -34.2\%                                    \\
                                                                                        & F + D + S              &                                                                           & 6.1                                 & 8.5             & 8.3               & 10.6           & 12.1          & 11.0               & 13.0           & 10.6                                         & -2.1\%                                        & -18.5\%                                    & -42.4\%                                    \\
                                                                                        & F + I                  &                                                                           & 9.8                                 & 9.6             & 9.6               & 10.4           & 11.3          & 10.6               & 12.1           & 10.6                                         & -38.9\%                                       & -18.5\%                                    & \textbf{-7.7\%}                            \\
                                                                                        & F + S + I              &                                                                           & 7.7                                 & 8.2             & 7.9               & 9.7            & 10.6          & 9.8                & 11.8           & 9.7                                          & -22.7\%                                       & -10.8\%                                    & -20.2\%                                    \\
                                                                                        & F + D + I              &                                                                           & 8.0                                 & 8.3             & 8.2               & 9.9            & 10.6          & 10.0               & 11.9           & 9.8                                          & -25.5\%                                       & -12.1\%                                    & -18.4\%                                    \\
                                                                                        & F + D + S + I          &                                                                           & 8.1                                 & 8.4             & 8.4               & 10.1           & 11.0          & 10.2               & 12.1           & 10.0                                         & -26.4\%                                       & -14.0\%                                    & -19.2\%                                    \\ \midrule
\multirow{8}{*}{\begin{tabular}[c]{@{}l@{}}Instance \\ segmentation (I)\end{tabular}}   & I                      & \multirow{8}{*}{\begin{tabular}[c]{@{}l@{}}mAP (\%) $\uparrow$, \\ vehicles\end{tabular}} & 63.9                                & 57.4            & 65.7              & 21.9           & 31.2          & 22.7               & 6.6            & 34.2                                         & -                                             & -                                          & 46.4\%                                     \\
                                                                                        & I + S                  &                                                                           & 64.9                                & 59.1            & 66.2              & 26.4           & \textbf{34.4} & 27.1               & \textbf{14.3}  & \textbf{37.9}                                & 1.5\%                                         & \textbf{10.7\%}                            & \textbf{41.6\%}                            \\
                                                                                        & I + S + D              &                                                                           & 65.0                                & 57.9            & 64.9              & 25.9           & 32.6          & 26.1               & 10.9           & 36.4                                         & 1.6\%                                         & 6.3\%                                      & 44.0\%                                     \\
                                                                                        & I + S + F              &                                                                           & 62.3                                & 57.1            & 64.2              & 21.6           & 31.6          & 23.6               & 7.7            & 34.3                                         & -2.5\%                                        & 0.1\%                                      & 45.0\%                                     \\
                                                                                        & I + D                  &                                                                           & \textbf{65.9}                       & \textbf{59.3}   & 66.9              & \textbf{26.8}  & 32.2          & \textbf{26.3}      & 11.4           & 37.2                                         & \textbf{3.1\%}                                & 8.5\%                                      & 43.6\%                                     \\
                                                                                        & I + D + F              &                                                                           & 65.8                                & 50.2            & \textbf{67.0}     & 21.5           & 31.0          & 22.9               & 6.7            & 33.2                                         & 2.9\%                                         & -3.0\%                                     & 49.5\%                                     \\
                                                                                        & I + F                  &                                                                           & 63.1                                & 56.9            & 65.1              & 20.4           & 28.9          & 21.7               & 5.0            & 33.0                                         & -1.3\%                                        & -3.6\%                                     & 47.7\%                                     \\
                                                                                        & I + S + D + F          &                                                                           & 64.8                                & 57.9            & 65.3              & 22.8           & 31.5          & 23.1               & 8.3            & 34.8                                         & 1.4\%                                         & 1.6\%                                      & 46.3\%                                     \\ \bottomrule
\end{tabular}
\caption{Multitask learning performances. We evaluate 15 combinations of 4 perception tasks: semantic segmentation (S), monocular depth estimation (D), optical flow estimation (F), and instance segmentation (I). The combinations of S + I, S + D, and S + D + I significantly improve on both tasks' source and OOD performance in their respective tasks. $\uparrow$ ($\downarrow$): the higher (lower) the better.}
\label{tab:multitask_four_tasks}
\end{table*}

\section{Additional Experiments} \label{sec:suppl_exp}
To further highlight the usefulness of \thedataset{},  we conduct experiments on multitask learning (\autoref{sec:suppl_exp_multitask}) and joint training with real-world data (\autoref{sec:suppl_exp_joint_training}). We also investigate the optimal dataset size and sampling rate (\autoref{sec:suppl_exp_dataset_size}).



\subsection{Multitask learning} \label{sec:suppl_exp_multitask}
In this experiment, we study whether different perception tasks mutually benefit or interfere with each other when jointly learned with a shared feature extractor.  
The wide variety of tasks supported in \thedataset{} unlocks new opportunities to  investigate different combinations of perception tasks. 
Special attention is also paid to the robustness of multitask models under incrementally shifted domains.


%

%

%
%

Specifically, we consider four different perception tasks: semantic segmentation, instance segmentation, monocular depth estimation, and optical flow estimation. 
%
%
Each task requires the model to learn a distinct encoding function: semantic segmentation requires intermediate activations to encode pixel-level information, instance segmentation requires instance-level information, depth estimation requires contextual information and object priors that allow to convert 2D images to 3D cues, and optical flow requires to encode a function of two images that embodies information on motion perception.

\paragraph{Multitask model.} To compose a unified multitask model, we use the segmentation model DRN-D-54~\citesupp{yu2017dilated} as feature extractor and combine it with the heads required for other tasks. The DRN-D-54 model has 8 sequential residual blocks with dilated convolutions and transposed convolutions at the end to generate segmentation results. 
Here, all the modules of DRN-D-54 are used for semantic segmentation. For instance segmentation, we rely on the Feature Pyramid Network (FPN)~\citesupp{lin2017feature}, Region Proposal Network (RPN), and ROIAlign modules identical to those introduce in Mask R-CNN~\citesupp{he2017mask}. 
FPN uses the 2nd to 5th blocks' outputs of the DRN-D network. For the optical flow and depth estimation, we adapt the decoders similar to FlowNet~\citesupp{dosovitskiy2015flownet} and U-Net~\citesupp{ronneberger2015u}. The decoder has 5 sequential blocks, where each block has one up-sampling layer, followed by a shortcut connection from the feature extractor's corresponding block, and a series of convolution layers. Together with the feature extractor, we obtain an encoder-decoder structure commonly used in dense prediction tasks.

\begin{figure*}[!t]
    \centering
    \includegraphics[width=1\linewidth]{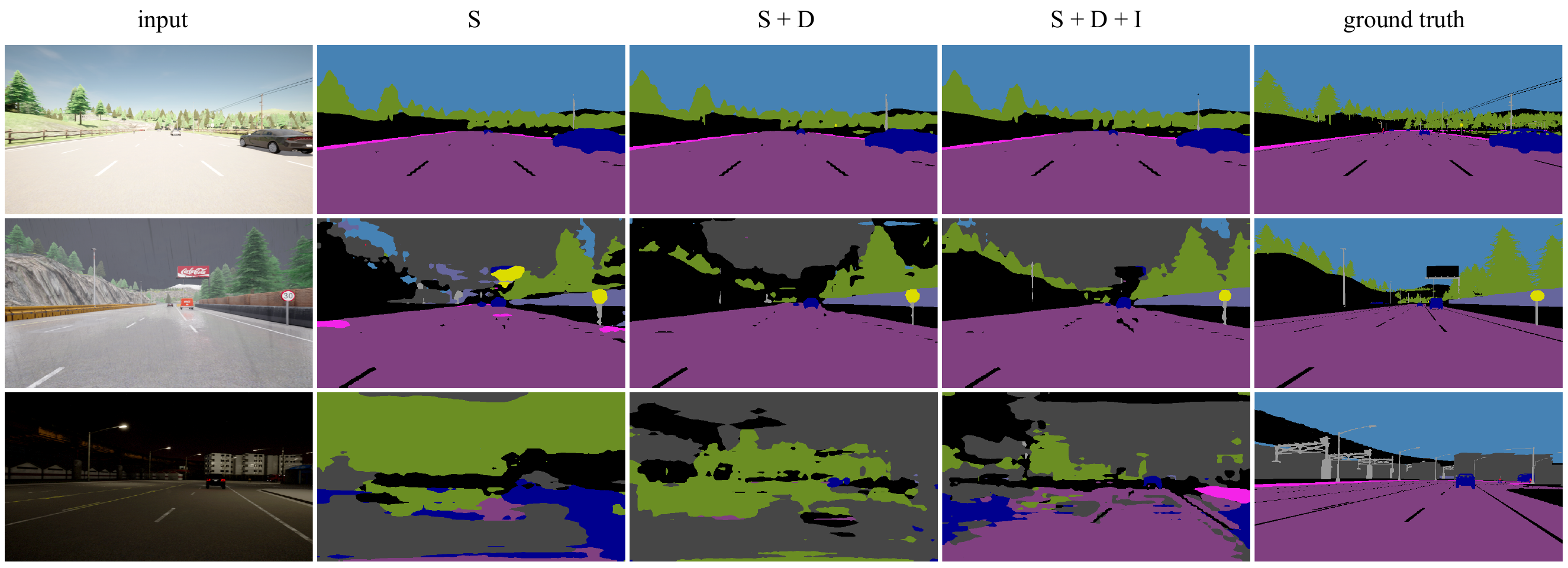}
    \caption{Qualitative results on semantic segmentation for models trained on the clear-daytime domain. Each column represents a model trained on a different task combination: semantic segmentation (S), semantic segmentation + depth estimation (S + D), semantic segmentation + depth estimation + instance segmentation (S + D + I). The three rows show the results on respectively the clear-daytime, rainy, and night domain. The combinations S + D and S + D + I improve the performance against domain shifts.}
    \label{fig:multitask_seg}
\end{figure*}

\paragraph{Experiment setup.} We traverse all 15 combinations for the 4 tasks mentioned above.
Our multitask model is trained with 5,000 frames sampled from the clear-daytime domain in \thedataset{} and evaluated under different discrete domain shifts. To fit the multitask model into the GPU memory, we reduce the image size to $640 \times 400$ pixels. 
Please note that the performance will be slightly affected by the size-reduced images and thus, it is not directly comparable to our baseline experiments in \ref{tab:ood_performance} of the main paper.
All combinations are trained for 100 epochs, when convergence is reached for all tasks. 

\paragraph{Experimental results} are summarized in \autoref{tab:multitask_four_tasks}. 
Every model is trained on the clear-daytime domain and tested on different types of shifted domains, indicated with OOD in the Table.
We report the average performance on the out-of-distribution domains as OOD avg.
The columns $\Delta_{\text{Source}}$   and $\Delta_{\text{OOD}}$  report for different multitask models the relative Source / OOD avg. performance change on a given task with respect to the performance of a single-task model trained on that specific task. 
The column $\Delta_{S~\xrightarrow{}~O.}$ reports for different multitask models the relative change from Source to OOD avg. performance on a given task. 
Below are our observations.

\textit{\textbf{Multitask learning improves robustness.}} 
We observe that specific combinations of tasks largely improve the single-task model performance on the source domain. For instance, the combination of semantic segmentation (S) + depth (D) + instance segmentation (I) boosts the source domain performance by 8.6\% / 11.0\% / 1.6\% on the respective tasks. 
Similar improvements are observed for other combinations, including S + I and S + D.  We visualize the results of these combinations in \autoref{fig:multitask_seg}. 
This is possibly due to the intertwined nature of such tasks. In particular, depth and semantics both need to learn contextual features from neighboring pixels, and both instance and semantics segmentation need to segment parts of the image.

Further, multitask learning often significantly increases the generalization of a model to domain shifts. For example, the combination of S + D + I improves the OOD performance in the respective tasks by 71.6\% / 42.8\% / 6.3\%. 
The improvements are substantially greater than the improvements on the source domain, suggesting that the increase in model's robustness is not attributable to the increase in the overall model's performance as seen on the source domain. 
We argue that this is potentially due to the model learning more general features that are shareable across tasks and, consequently, also more general under domain shifts. 
For example, the addition of instance segmentation typically causes the greatest robustness improvements. This might be due to the complex nature of the instance segmentation task, which requires to encode features capable of both detecting and segmenting objects in an image.
%


\textit{\textbf{Instance segmentation can only be improved mildly.}} Instance segmentation is only improved at most by 10.7\% on OOD performance by other tasks. 
As previously mentioned, we hypothesize that instance segmentation already learns more general features due to its nature. Thus, the addition of other tasks provides only mild improvements.
On the other hand, however, when combined with other tasks, instance segmentation largely boosts their robustness, \eg{} S + I and D + I.

\textit{\textbf{Optical flow is heavily affected by other tasks.}} Unlike the previous tasks that benefit from multitask learning, optical flow shows a different behavior.  
Although optical flow can improve other tasks' robustness (\eg{} S + F and D + F), the optical flow itself is negatively affected by the addition of other tasks. 
When jointly trained with other tasks, its performance drops by a large margin, ranging from -0.2\% to -38.9\%. 
A possible explanation is that the optical flow task, which takes a pair of frames as input, learns a different encoding function than other non-temporal tasks requiring only one frame.
To learn a feature extractor shared across the two different types of inputs, the model shows to sacrifice its effectiveness on the task requiring two images.
This suggests that combining different tasks is not trivial; instead, it requires extensive evaluation and comparison.
\thedataset{} provides a playground to develop novel multitask learning techniques and to investigate and solve the multiple challenges presented by such an interesting problem. 


\textit{\textbf{Domain shift is only partially mitigated.}}  
While the model's robustness can be improved by multitask learning, the domain shifts provided in \thedataset{} still pose a tremendous threat to the robustness under domain shift. 
For all the evaluated tasks, the minimum average OOD performance drop with respect to the corresponding source performance ($\Delta_{S~\xrightarrow{}~O.}$) amounts to $\sim$~40\%. 
Under extreme conditions, \eg{} foggy and night, the performances are degraded even more than 60\%, which indicates real-life risks if autonomous vehicles heavily rely on such models.

By introducing \thedataset{}, which supports multi-domain and multitask studies in a single dataset, we hope to foster future research on multitask domain adaptation algorithms to counteract these domain gaps effectively.
Moreover, we hope that the continuous domain shifts provided in our dataset will shed new light on this challenging problem.

\subsection{Joint training with real-world data} \label{sec:suppl_exp_joint_training}

We investigate whether the domain variations in our dataset in combination with a specific domain of real-world data can make a model more robust to domain shift compared to a model only trained on the real-world data. 
Specifically, we jointly train the model with the source domain data (\textit{i.e.}, clear daytime) from BDD100K and all domain variations from ours. The model is then evaluated on other domains of BDD100K.
We employ the Faster R-CNN~\citesupp{ren2015faster} as the model for object detection and DRN-D-54~\citesupp{yu2017dilated} for semantic segmentation. The models are learned with the same amount of data from BDD100K but with different amounts of data from \thedataset{}. 

\paragraph{Object detection} results are shown in \autoref{tab:det_joint}. We observe that the joint training provides a relative improvement of the source domain and OOD performance amounting to 2.52\% and 3.40\%, respectively.

\paragraph{Semantic segmentation} has  similar trends. As shown in \autoref{tab:seg_joint}, source domain mIoU improves from 46.04\% to 51.20\%, with a relative improvement of 10.34\%. Moreover, out-of-domain mIoU rises by a relative 5.30\% from 37.37\% to 39.76\%. 

These results suggest that, if a model is trained on a limited real-world domain, jointly training with the variety of domains provided by our dataset will improve the robustness of the model to real-world shifts.


\begin{table}[h]
\centering
\footnotesize
\setlength{\tabcolsep}{8pt}
\begin{tabular}{lcccc}
\toprule
\multirow{2}{*}{\textbf{Training set}} & \multicolumn{2}{c}{\textbf{Source domain}} & \multicolumn{2}{c}{\textbf{OOD avg.}}    \\ 
\cmidrule(lr){2-3} \cmidrule(lr){4-5}
                            & \multicolumn{1}{c}{AP} & AP$_{75}$ & AP    & AP$_{75}$ \\ \midrule
BDD100K                     & 0.318                  & 0.312     & 0.265 & 0.251     \\
BDD100K + 2k frames  & 0.320                   & 0.327     & 0.267 & 0.267     \\
BDD100K + 5k frames    & \textbf{0.326} & \textbf{0.334} & \textbf{0.274} & \textbf{0.271} \\
BDD100K + 10k frames & 0.325                  & 0.329     & 0.254 & 0.238     \\
\bottomrule
\end{tabular}
\caption{Joint training for object detection. Generalization ability is improved with a proper amount of data. \label{tab:det_joint}}
\end{table}
\begin{table}[h]
\centering
\footnotesize
\setlength{\tabcolsep}{8pt}

\begin{tabular}{lcc}
\toprule
 \textbf{Training set}        &  \textbf{Source domain}   & \textbf{OOD avg.} \\ \midrule
BDD100K                     & 46.04   & 37.37                             \\
BDD100K + 6k frames  & 47.11 &  38.56                             \\
BDD100K + 12k frames &
  \textbf{51.20} &
  \textbf{39.76} \\
BDD100K + 24k frames & 51.09 & 39.23   \\
\bottomrule
\end{tabular}
\caption{Joint training for semantic segmentation. We report the mIoU. Generalization ability is improved with a proper amount of data. \label{tab:seg_joint}}
\end{table}

\subsection{Dataset size} \label{sec:suppl_exp_dataset_size}
To understand the impact of dataset size and optimize the design of the dataset, we conduct ablation studies on: (1) sampling rate and (2) amount of sequences. 
Every model is trained on clear-daytime sequences.


\begin{table}[t]
    \centering
    \footnotesize
    \setlength{\tabcolsep}{5pt}
    \begin{tabular}{l|cccccc}
    \toprule
       \textbf{Frame rate (Hz)}  & 0.1 & 0.2 & 0.5 & 1 & 5 & 10 \\
       \midrule
       \textbf{\# Frames ($\times$1k)} & 7.5 & 15 & 37.5 & 75 & 375 & 750 \\ \midrule
        Seg. (mIoU, \%) & 62.6 & 62.9 & \textbf{63.1} & 63.0 & 62.9 & - \\
        Det. (mAP, \%) & 40.6 & 43.1 & 45.8  & 46.8 & \textbf{48.4} & - \\
        MOT (MOTA, \%) & 25.6  & 34.7 & 45.2 & 49.3 & 54.1 &  \textbf{54.9} \\
        \bottomrule
    \end{tabular}
    \caption{\footnotesize{Performance of different tasks at increasing sampling rates. Training and testing on the same 1500 sequences from all domains.}}
    \label{tab:frame_rate}
    \vspace{-1em}
\end{table}

\begin{table}[h]
    \centering
    \footnotesize
    \setlength{\tabcolsep}{5pt}
    \begin{tabular}{l|cccccc}
    \toprule
       \textbf{Training sequence }  & 350 & 750 & 1500 & 2000 & 3000  \\
       \midrule
        Seg. (mIoU, \%) & 59.4 & 61.4 & 63.0 & 62.6 & \textbf{63.1}\\
        Det. (mAP, \%) & 41.2 & 45.1 & 46.8  & 48.0 & \textbf{50.1} \\
        \bottomrule
    \end{tabular}
    \caption{\footnotesize{Performance of different tasks at increasing sequences number. Training and testing on the data of 1Hz from all domains.}}
    \label{tab:sequences}
    \vspace{-1em}
\end{table}

\paragraph{Frame rate.} 
%
To avoid the model learning from redundant information, we study what is the optimal sampling rate to achieve the best performance on a given task.
Here, we test the semantic segmentation, object detection, and multiple obeject tracking performance on a set of images sampled at different frame rates from a fixed set of 2000 sequences. 
We notice that performance of different tasks starts to saturate at different sampling rates  (\autoref{tab:frame_rate}). 
For image-based tasks, such as segmentation and detection, we argue that the information provided by adjacent frames can be redundant, and increasing the sampling rate over a certain threshold have insignificant benefits on the resulting model performance.
However, for video-based tasks, like multi-object tracking, the inter-frame information is crucial. A lower frame rate leads to lose a considerable amount of information, thus severely reducing the model performance (\autoref{tab:frame_rate}, third row). 

Our dataset is collected at a fixed frame rate of 10Hz, which is necessary to support a wide range of perception tasks.  
However, according to the experiments on the sampling rate, we also provide a subset sampled at 1Hz for image-based perception tasks.

%
%

\paragraph{Amount of sequences} is another factor affecting the performance. 
Here, we test semantic segmentation and object detection performance on a varying number of sequences sampled at 1Hz.
In \autoref{tab:sequences}, we find that the performance continuously increases up to 3000 sequences. 
However, the performance gain is diminishing the more sequences we add. This is potentially due to the limited environmental variation in the simulator. 
To balance between size and learning performance, we set the total number of sequences to 3000 for the discrete set. 
Together with our sampling pipeline (\autoref{ssec:dataset_generation}), the current size of \thedataset{} guarantees 
that for each BDD100K's domain label, we have more than 500 corresponding sequences for training and testing.

\subsection{Comparison with VIPER} \label{sec:viper}
As a synthetic dataset, VIPER~\cite{Richter_2017} also presents sequences from discrete domain shifts. 
Here, we compare the segmentation performance under domain shifts in VIPER, SHIFT and BDD100K (\autoref{tab:viper}). We find that the adverse conditions presented in VIPER provide a less relevant threat to model generalization, highlighting how SHIFT mimics more closely real-world trends.

\begin{table}[!ht]
    \centering
    \footnotesize
    \setlength{\tabcolsep}{5pt}
    \begin{tabular}{l|cccc|c}
    \toprule
       \textbf{Dataset} & \underline{daytime} ($M_0$) & sunset & night & rain &  $\frac{\max{\Delta M}}{M_{0}}$ \\
       \midrule
        VIPER & 59.3 & 57.6 & 55.1 & 53.0  & -10.6\% \\
        SHIFT (\textit{ours})  & 83.6 & 60.4 & 42.8 & 54.6 & -48.8\% \\ \midrule
        BDD100K & 47.9 & - & 20.6 & 37.6  &  -57.0\% \\
        \bottomrule
    \end{tabular}
    \caption{\footnotesize{Out-of-distribution performance on different datasets of a segmentation model (DRN-D) trained on the daytime domain. The last column represents the maximal relative performance drop w.r.t. source.}}
    \label{tab:viper}
    \vspace{-1em}
\end{table}

\subsection{Error analysis for foggy and rainy domains} \label{sec:error}
As noticeable in \autoref{fig:trend}, detection and segmentation models show a slightly different behavior under different types of domain shift. While it is worth noticing that segmentation and detection have different label sets, we here analyze the differences in performance on the two domains presenting the largest discrepancy across the two tasks, \ie{} foggy and rainy. 
%
%
For example, we find that the most drastically affected class for segmentation in the rainy domain is `sky' (-69\% mIoU w.r.t. clear-daytime), with 25\% of the corresponding pixels misclassified as `building', as opposed to only 2\% under foggy conditions. 
%
%
For object detection, we find that most of the errors come from missed detections. The shifted domains lower the classification confidence below the pre-selected threshold, with foggy posing a greater challenge (car AP drops by 74\% on foggy vs 40\% on rainy).

\section{Implementation Details} \label{sec:implementation_details}
In this section, we describe the implementation details and metrics for each task in \autoref{tab:ood_performance} and \autoref{fig:trend}.

\paragraph{Object detection.} We compare Faster R-CNN~\cite{ren2015faster}, Cascade R-CNN~\cite{cai2018cascade}, and YOLO v3~\cite{redmon2018yolov3}. The backbone network for the first two methods is ResNet-50~\cite{he2016deep}, while YOLO v3 uses DarkNet~\cite{redmon2017yolo9000} as its backbone. We use the mean Average Precision (mAP) as the metric for 2D bounding boxes. We train the models on 50k frames of data, following the ``1x'' schedule provided in the \texttt{mmdetection} library~\citesupp{mmdetection}.

\paragraph{Semantic segmentation.} We also compare three models for semantic segmentation, DeepLab v3+~\cite{deeplabv3plus2018}, Fully Convolutional Network (FCN)~\cite{long2015fully}, and DRN-D-54~\cite{yu2017dilated}. All three models use the ResNet-50~\cite{he2016deep} as the backbone. We train the models with 20k frames of data until they converge (approximately 150 epochs). We use the mean IoU (mIoU) metric for all evaluations on semantic segmentation.

\paragraph{Instance segmentation.} We use Mask R-CNN~\cite{he2017mask} with ResNet-50 backbone and follow the same training routine as  Faster R-CNN~\cite{ren2015faster}. A segmentation mAP metric is used for evaluation.

\paragraph{Depth estimation.} We use AdaBins~\cite{bhat2021adabins} for the depth estimation experiments. It uses a U-Net-like~\cite{ronneberger2015u} backbone structure and predicts depth with adaptive bins. The model is trained using its official implementation. We follow KITTI's benchmark on depth estimation~\cite{geiger2013vision}. Specifically, we use the Scale-invariant Logarithm (SILog) metric evaluated on the central crop of the image (\ie{} Eigen Crop).

\paragraph{Optical flow estimation.} We use RAFT~\cite{teed2020raft} for optical flow estimation. The model is fine-tuned from pre-trained weights on the Things Dataset~\cite{mayer2016large}, with 10k frames of our data. The End-point Error (EPE) metric is used for evaluation.


{\small
\bibliographystylesupp{ieee_fullname}
\bibliographysupp{egbib}
}

\end{document}